%% file: main.tex
\PassOptionsToPackage{table}{xcolor}
\documentclass[10pt,twocolumn,letterpaper]{article}
\usepackage[T1]{fontenc}

\usepackage{cvpr}

\definecolor{cvprblue}{rgb}{0.21,0.49,0.74}
\usepackage[pagebackref,breaklinks,colorlinks,allcolors=cvprblue]{hyperref}
\usepackage{multirow}
\usepackage{pifont}

\def\confName{CVPR}
\def\confYear{2027}

\title{Rethinking Visual Token Compression for Video Large Language Models:\\A Simple Yet Strong Baseline}

\author{
Xiao Zhang$^{1,2}$,
Wang Zeng$^2$,
Sheng Jin$^{2*}$,
Yunzhe Liu$^1$,
Wentao Liu$^2$,
Chen Qian$^2$,
Shichao Kan$^{1\dagger}$\\
$^1$School of Computer Science and Engineering, Central South University\\
$^2$SenseTime Research and Tetras.AI\\
{\tt\small
xiaozhang@csu.edu.cn \quad
jinsheng@tetras.ai \quad
kanshichao@csu.edu.cn
}
}

\begin{document}
\maketitle

\begingroup
\renewcommand{\thefootnote}{}
\footnotetext{\textsuperscript{*}Project leader. \quad
\textsuperscript{$\dagger$}Corresponding author.}
\endgroup

\input{secs/0_abstract}
\input{secs/1_intro}
\input{secs/2_related}
\input{secs/3_method}
\input{secs/4_experiments}
\input{secs/5_analysis}
\input{secs/6_conclusion}
\input{secs/7_limitation}

\section*{LLM Usage Statement}
In this work, we used Large Language Models (LLMs) to improve the clarity, grammar, and readability of the paper. After using these tools, we carefully reviewed and edited the content as needed and take full responsibility for the final content of this work.

{
\small
\bibliographystyle{ieeenat_fullname}
\bibliography{main}
}

\input{secs/X_suppl}

\end{document}

%% file: secs/0_abstract.tex
\begin{abstract}
Video large language models (Video LLMs) have achieved remarkable progress in video understanding, but their inference efficiency is constrained by the large number of visual tokens produced by long videos. Recent video token compression methods increasingly rely on complex, hand-crafted heuristics for token pruning or merging. This raises a fundamental question: how much of compression performance can be obtained by simply preserving the structure encoded in the visual representations? We investigate this question with SimpleComp, a simple and training-free baseline that performs position-aware cross-frame clustering in the visual feature space and represents each cluster using the mean of its original visual features. Extensive experiments across five video understanding benchmarks and three representative Video LLMs show that SimpleComp outperforms recent compression methods at most visual token retention ratios, with particularly strong robustness under extremely low retention rates. To understand this behavior, we analyze the feature space preserved by different compression methods in terms of local approximation fidelity and global coverage. The results show that stronger downstream performance is consistently associated with better preservation of the original visual feature distribution, especially its global coverage. These findings highlight feature-space preservation as an important consideration for video token compression under highly constrained token budgets. Our code is available at \url{https://github.com/xiaozhang79/SimpleComp}.
\end{abstract}

%% file: secs/1_intro.tex
\section{Introduction}
\label{sec:introduction}

Video large language models (Video LLMs) have achieved remarkable capabilities in video understanding by combining powerful vision encoders with large language models~\cite{li2025llava,zhang2025llava,zhu2025internvl3,bai2025qwen2,jin2023chatunivi}. However, processing long videos remains computationally expensive due to the large number of visual tokens produced by dense spatial-temporal representations. This has motivated a growing body of research on visual token compression, which aims to reduce the visual sequence length while preserving downstream understanding performance~\cite{yang2025visionzip,shen2025fastvid,tao2025dycoke,wang2026earlytom,li2026token,shao2026survey}.

Early studies on visual token compression mainly focused on image understanding, exploring token merging, pruning, and selection to reduce visual redundancy~\cite{bolya2022token,chen2024image,xing2025conical}. Recent methods extend this paradigm to Video LLMs by reducing visual redundancy through importance-based selection, density-aware pruning, token merging, context-aware aggregation, and spatiotemporal compression~\cite{yang2025visionzip,shen2025fastvid,shao2025holitom,wang2026earlytom,li2026token,shen2025longvu,wang2026streaming}. These approaches provide increasingly sophisticated mechanisms for deciding which visual tokens to retain or merge. However, it remains unclear how much of the compression performance depends on such elaborate selection mechanisms, and how much can be obtained by simply preserving the structure already encoded in the visual representations.

\begin{figure*}[t]
\centering
\includegraphics[width=\textwidth]{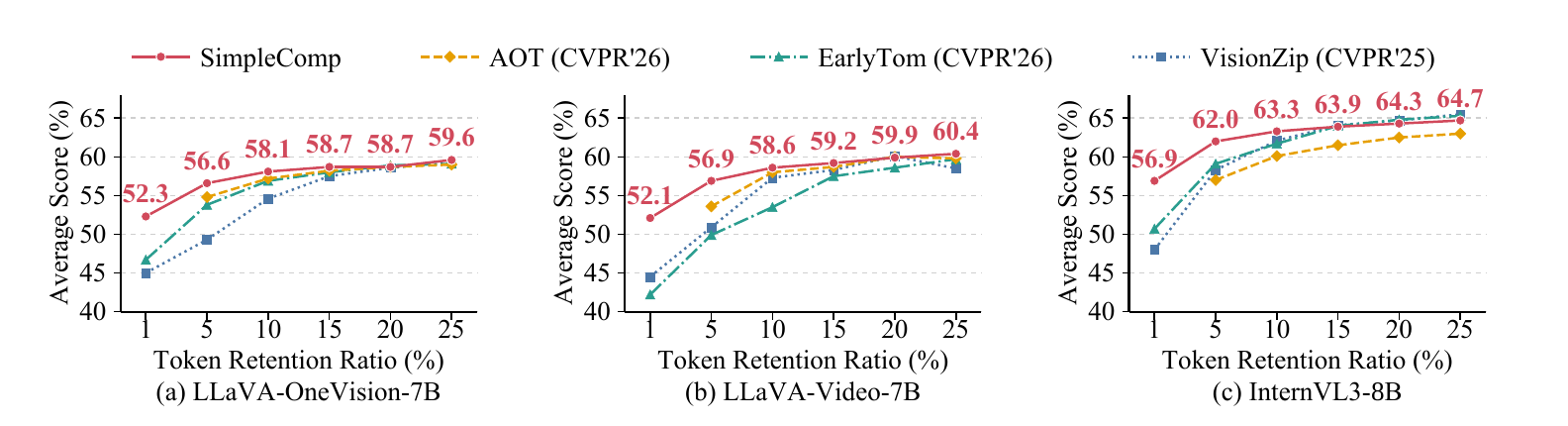}
\caption{Performance comparison of SimpleComp with recent visual token compression methods across different visual token retention ratios on LLaVA-OneVision-7B, LLaVA-Video-7B, and InternVL3-8B. SimpleComp outperforms existing methods in most settings, particularly at extremely low token retention ratios.}
\label{fig:cmp}
\end{figure*}

We explore this question with \textit{SimpleComp}, a deliberately simple and training-free baseline for video token compression. SimpleComp operates on visual tokens after the projector and performs cross-frame clustering in a position-aware feature space. Each cluster is represented by the mean of its original visual features, and the resulting compact sequence is directly fed into the LLM. The method introduces no additional training or architectural modification. Despite this simple design, SimpleComp outperforms recently published compression methods at most visual token retention ratios across multiple Video LLMs, with particularly strong robustness under extremely low retention rates, as illustrated in Figure~\ref{fig:cmp}.

The strong performance of such a simple baseline raises a more fundamental question: \emph{what makes feature-space clustering effective for visual token compression?} We investigate this question from the perspective of feature-space preservation. Since SimpleComp directly approximates the visual feature distribution with a limited set of representatives, we analyze both local approximation fidelity and global coverage of the original feature space. Across different compression methods and token budgets, we find that stronger downstream performance is consistently associated with better preservation of the original visual feature distribution. In particular, SimpleComp maintains substantially broader feature-space coverage while preserving local feature fidelity, providing an empirical explanation for its robustness under aggressive compression.

Our contributions are summarized as follows:
\begin{itemize}
\item We introduce \textit{SimpleComp}, a simple and training-free baseline for video token compression that performs position-aware cross-frame clustering in the visual feature space, without additional training or architectural modification.

\item Through extensive experiments on three representative Video LLMs and five video understanding benchmarks, we show that SimpleComp outperforms existing compression methods at most visual token retention ratios and exhibits strong robustness under aggressive compression.

\item We analyze why simple feature-space clustering remains effective under constrained token budgets and find that stronger compression performance is consistently associated with better preservation of the original visual feature distribution, especially its global coverage. This observation highlights feature-space preservation as an important consideration for video token compression.
\end{itemize}

%% file: secs/2_related.tex
\section{Related Work}
\label{sec:related_work}

\subsection{Video Large Language Models}

Recent advances in multimodal large language models (MLLMs) have substantially improved general visual-language understanding, with Video LLMs extending these capabilities to video perception and reasoning~\cite{li2025llava,zhang2025llava,zhu2025internvl3,bai2025qwen2,bai2025qwen3,jin2023chatunivi,li2024videochat,maaz2024videochatgpt,wang2024videollama2,zhang2026thinking}. Representative Video LLMs include LLaVA-OneVision~\cite{li2025llava} unifies single-image, multi-image, and video understanding through visual task transfer. LLaVA-Video~\cite{zhang2025llava} improves video instruction following through large-scale synthetic video-language data, while InternVL3~\cite{zhu2025internvl3} adopts native multimodal pre-training to strengthen general visual reasoning.

Despite these advances, Video LLMs remain computationally expensive when processing long videos, as dense spatiotemporal representations generate a large number of visual tokens. This has motivated growing interest in visual token compression, which aims to reduce the visual sequence length while preserving downstream video understanding performance.

\subsection{Visual Token Compression}

Visual token compression has been widely explored to improve the efficiency of MLLMs. Early studies mainly focused on image understanding, exploring token merging, pruning, and importance-based selection to reduce redundant visual representations~\cite{bolya2022token,chen2024image,xing2025conical,alvar2025divprune}.

Recent methods extend this paradigm to Video LLMs by further exploiting spatial and temporal redundancy across video frames~\cite{yang2025visionzip,tao2025dycoke,shen2025fastvid,shao2025holitom,wang2026earlytom,li2026token,fu2024framefusion,huang2024prunevid,shao2026survey}. VisionZip~\cite{yang2025visionzip} selects dominant visual tokens based on attention and merges redundant ones. FastVID~\cite{shen2025fastvid} combines temporal segmentation with density-aware pruning. EarlyTom~\cite{wang2026earlytom} performs early-stage token compression within the vision encoder. AOT~\cite{li2026token} introduces attention-guided anchors and optimal transport for token aggregation.

Despite their strong performance, these methods typically rely on increasingly complex, hand-crafted designs, raising the question of whether such elaborate designs are necessary. We explore this question with SimpleComp, a deliberately simple and training-free baseline for video token compression that performs position-aware cross-frame clustering in the visual feature space and outperforms recent compression methods at most visual token retention ratios, particularly under aggressive compression.

%% file: secs/3_method.tex
\begin{figure*}[t] 
\centering
\includegraphics[width=0.9\textwidth]{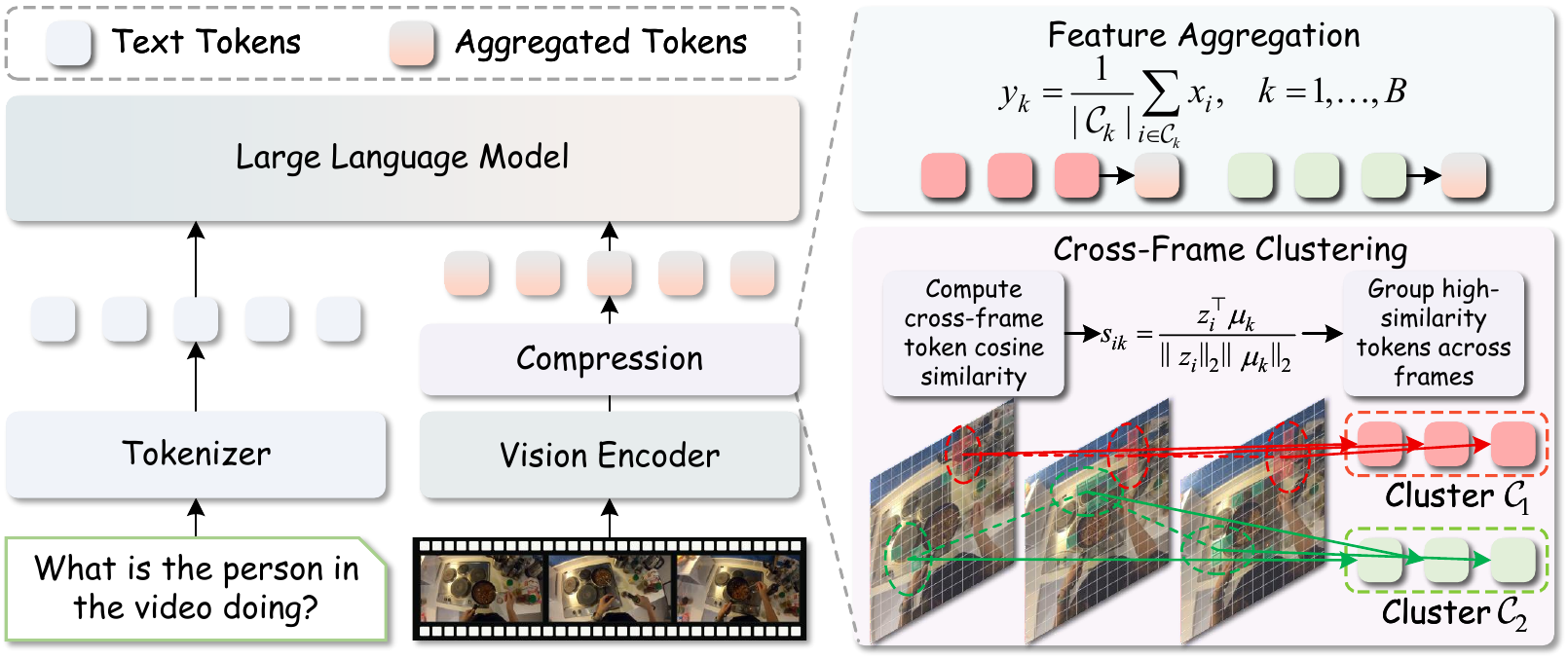}
\caption{Overall pipeline of SimpleComp. SimpleComp compresses visual tokens through position-aware cross-frame clustering in the visual feature space, with each cluster represented by the mean of its original visual features.}
\label{fig:overview}
\end{figure*}

\section{From Complex Video Token Compression to SimpleComp}
\label{sec:method}

\subsection{Problem Formulation}
Given an input video $\mathcal{V}$ and a text query $\mathcal{Q}$, a Video LLM processes $T$ sampled frames to produce a sequence of projected visual tokens $X=[x_1,\ldots,x_N]\in\mathbb{R}^{N\times d}$, where $N$ and $d$ denote the number of visual tokens and feature dimension, respectively. Video token compression aims to transform this dense sequence into a compact representation $Y=[y_1,\ldots,y_B]\in\mathbb{R}^{B\times d}$, where $B\ll N$ denotes the token budget and $r=B/N$ is the token retention ratio. The goal is to preserve the original feature-space structure under a substantially reduced token budget.

Existing methods typically employ complex, hand-crafted heuristics, such as attention-based selection, density-aware pruning, and spatiotemporal redundancy modeling, to guide token compression. Differently, we introduce SimpleComp, a deliberately simple baseline that performs position-aware cross-frame clustering in the visual feature space and represents each cluster using the mean of its original visual features, as shown in Figure~\ref{fig:overview}.

\subsection{SimpleComp: A Simple Feature-Space Clustering Baseline}
\label{sec:simplecomp}
\noindent \textbf{Position-Aware Feature Construction.}
Visual similarity alone does not fully characterize the structure of video tokens, as tokens with similar visual features may appear at different spatial or temporal positions. We therefore incorporate spatiotemporal positional information into the features used for clustering. For each visual token $x_i$, we denote its position by $p_i=(f_i,h_i,w_i)$, where $f_i$, $h_i$, and $w_i$ correspond to its frame, row, and column indices, respectively. We apply 3D Rotary Position Embedding (RoPE)~\cite{su2024roformer} and normalize the resulting feature:
\begin{equation}
z_i =
\frac{R(p_i)x_i}
{\|R(p_i)x_i\|_2},
\end{equation}
where $R(p_i)$ denotes the rotary transformation associated with position $p_i$. The resulting position-aware features are used exclusively for clustering, while the original projected features $x_i$ are retained for subsequent aggregation. Importantly, this additional 3D RoPE operates independently of the backbone's native positional encoding. For models such as Qwen3-VL~\cite{bai2025qwen3}, which employ interleaved M-RoPE within the LLM to encode multimodal positions, our 3D RoPE serves a separate purpose: incorporating spatiotemporal information into feature similarity during clustering, without modifying the backbone's positional encoding.

\noindent \textbf{Cross-Frame Clustering.}
Given the position-aware features $\{z_i\}_{i=1}^{N}$, we jointly cluster visual tokens from all $T$ sampled frames into $B$ clusters. Specifically, we employ an iterative cosine-similarity-based clustering procedure. At each iteration, every token is assigned to the cluster with the most similar prototype:
\begin{equation}
c_i =
\arg\max_{k\in\{1,\ldots,B\}}
\frac{z_i^\top\mu_k}
{\|z_i\|_2\|\mu_k\|_2},
\end{equation}
where $\mu_k$ denotes the prototype of the $k$-th cluster and $c_i$ is the cluster assignment of the $i$-th token. The prototypes are subsequently updated by averaging their assigned position-aware features. These assignment and update steps are repeated until the cluster assignments converge. Joint clustering across frames exploits cross-frame redundancy and enables adaptive allocation of the limited token budget according to the global visual feature distribution.

Since a cluster may contain tokens from multiple frames, we assign its spatiotemporal position using the earliest token in that cluster. Specifically, let $i_k^*$ denote the member with the earliest frame index in cluster $\mathcal{C}_k$, with spatial coordinates used to break ties. The corresponding position $p_{i_k^*}$ serves as the cluster's representative position. We then order the resulting clusters according to their representative positions to construct the compressed visual token sequence.

\noindent \textbf{Cluster-Wise Feature Aggregation.}
After obtaining the cluster assignments, we aggregate the original projected visual features within each cluster using mean pooling to construct compact visual tokens:
\begin{equation}
y_k =
\frac{1}{|\mathcal{C}_k|}
\sum_{i\in\mathcal{C}_k} x_i,
\quad
k=1,\ldots,B,
\end{equation}
where $\mathcal{C}_k=\{i\mid c_i=k\}$ denotes the tokens assigned to the $k$-th cluster. Note that aggregation is performed on the original projected features rather than the position-aware features used for clustering. The resulting $B$ aggregated visual tokens, together with the text tokens, are fed into the large language model for downstream video understanding.

%% file: secs/4_experiments.tex
\section{Experiments}
\label{sec:experiments}

\subsection{Experimental Setup}
\noindent \textbf{Benchmarks and Compared Baselines.}
We evaluate our SimpleComp on five widely used video understanding benchmarks, including VideoMME~\cite{fu2025video}, LongVideoBench~\cite{wu2024longvideobench}, MLVU~\cite{zhou2025mlvu}, MVBench~\cite{li2024mvbench}, and EgoSchema~\cite{mangalam2023egoschema}. These benchmarks cover diverse video durations and scenarios for comprehensive evaluation of visual token compression. We compare our method with several recent representative video token compression methods, including FastVID~\cite{shen2025fastvid}, VisionZip~\cite{yang2025visionzip}, EarlyTom~\cite{wang2026earlytom}, and AOT~\cite{li2026token}.

\noindent \textbf{Implementation Details.}
We conduct comprehensive experiments at 1\%, 5\%, 10\%, 15\%, 20\%, and 25\% token retention ratios on three representative Video LLMs: LLaVA-OneVision-7B~\cite{li2025llava}, LLaVA-Video-7B~\cite{zhang2025llava}, and InternVL3-8B~\cite{zhu2025internvl3}. For LLaVA-OneVision-7B and LLaVA-Video-7B, we adopt available baseline results reported in EarlyTom~\cite{wang2026earlytom} and AOT~\cite{li2026token}, and further evaluate the remaining settings using their released implementations. For InternVL3-8B, we extend the released baseline implementations to this backbone while following their original compression procedures and hyperparameter settings. Following the official setting, LLaVA-OneVision, LLaVA-Video, and InternVL3 uniformly sample 32, 64, and 32 frames, respectively. During clustering, we adopt deterministic farthest-point initialization and perform at most 10 iterations. All experiments are conducted on NVIDIA A800 80GB GPUs using LMMs-Eval~\cite{zhang2025lmms}.

\begin{figure}[t]
\centering
\includegraphics[width=\columnwidth]{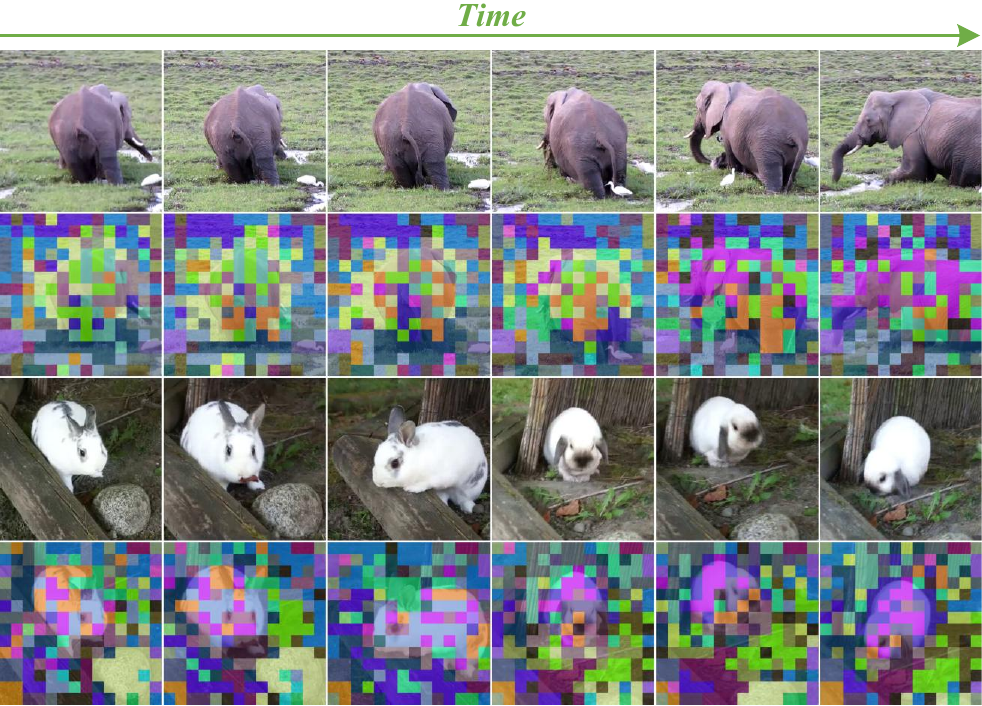}
\caption{Cross-frame video clustering visualizations generated by SimpleComp. The same color indicates the same cluster.}
\label{fig:vis}
\end{figure}

\input{tabs/llava_ov}
\input{tabs/llava_video}
\input{tabs/internvl3}
\input{tabs/diff_baselines}

\subsection{Main Results}
Tables~\ref{tab:llava_ov}, \ref{tab:llava_video}, and \ref{tab:internvl3} compare our SimpleComp with recent visual token compression methods across different retention ratios on three Video LLM backbones. Overall, SimpleComp achieves consistently strong performance, with its advantages becoming more pronounced as the token budget decreases. At a 15\% retention ratio, SimpleComp achieves the best average scores on LLaVA-OneVision-7B and LLaVA-Video-7B, while remaining competitive on InternVL3-8B. At 10\%, SimpleComp outperforms all competing methods across all three backbones, retaining 97.9\%, 94.8\%, and 96.1\% of the corresponding uncompressed model performance, respectively.

As the retention ratio decreases to 5\%, SimpleComp achieves average scores of 56.6, 56.9, and 62.0, outperforming the strongest baselines by 1.8, 3.3, and 2.9 points, respectively. Even at an extremely low retention ratio of 1\%, SimpleComp retains 88.2\%, 84.3\%, and 86.4\% of the uncompressed performance, surpassing the strongest baselines by 5.6, 7.6, and 6.2 points. These results demonstrate the robustness of SimpleComp under aggressive visual token compression, with increasingly substantial advantages at lower token budgets. Additional results at 20\% and 25\% retention ratios are provided in the supplementary material.

To further illustrate how SimpleComp exploits cross-frame visual redundancy, Figure~\ref{fig:vis} presents the clustering results for two video sequences, where tokens assigned to the same cluster are represented by the same color. As shown, visually similar tokens from different frames are grouped into shared clusters, even when their spatial positions and appearances change over time. This demonstrates that SimpleComp effectively captures feature similarities across frames, enabling redundant visual information to be consolidated into compact representations while preserving diverse visual content.

\subsection{Comparison with Simple Baselines}
To examine the effectiveness of feature-space clustering, we compare SimpleComp with several simple visual token compression baselines, including Spatial Average Pooling, Spatiotemporal Average Pooling, Spatial Grid Sampling, and Spatiotemporal Grid Sampling. For grid-based sampling, tokens are consistently selected from the top-left regions of the spatial or spatiotemporal grids. All methods are evaluated on LLaVA-OneVision-7B.

As shown in Table~\ref{tab:diff_baselines}, SimpleComp achieves the highest average scores of 58.1, 56.6, and 52.3 at 10\%, 5\%, and 1\% retention ratios, respectively, consistently outperforming all competing strategies. While these simple strategies reduce visual tokens through sampling or pooling operations, they do not explicitly consider the underlying visual feature structure. In contrast, SimpleComp directly groups visually similar tokens across frames and aggregates their original features into compact representations. These results demonstrate the effectiveness of feature-space clustering for video token compression, particularly under extremely low token budgets. Additional results at 15\%, 20\%, and 25\% visual token retention ratios are provided in the supplementary material.

 \input{tabs/ablations}

\subsection{Ablation Studies}
We examine the key design choices of SimpleComp, including positional encoding, clustering scope, and cluster representation strategies. All ablation experiments are conducted on LLaVA-OneVision-7B at a 10\% token retention ratio.

\noindent \textbf{Effect of Positional Encoding.}
We first examine the effect of incorporating positional information into feature-space clustering through 3D RoPE. As shown in Table~\ref{tab:ablation}, removing 3D RoPE reduces the average score from 58.1 to 57.3. This indicates that incorporating spatiotemporal positional information into the clustering space helps distinguish visually similar tokens at different spatial or temporal positions, leading to more effective token grouping.

\noindent \textbf{Effect of Clustering Scope.}
We next compare cross-frame clustering with per-frame clustering. As shown in Table~\ref{tab:ablation}, cross-frame clustering improves the average score from 57.0 to 58.1. This demonstrates the benefit of jointly clustering visual tokens across frames, which allows better exploitation of cross-frame redundancy and adaptive allocation of the limited token budget according to the global feature distribution.

\noindent \textbf{Effect of Cluster Representation Strategies.}
Finally, we compare two strategies for representing each cluster: selecting the nearest token and aggregating the original visual features using their mean. As shown in Table~\ref{tab:ablation}, both strategies achieve comparable average scores of 58.0 and 58.1, respectively. Nearest-token selection performs better on VideoMME and MVBench, while mean aggregation achieves higher scores on the other three benchmarks. These results indicate that both representation strategies provide comparable overall performance, with mean aggregation offering a slight advantage in average performance.

\input{tabs/efficiency}

\subsection{Efficiency Analysis}
Although visual token compression reduces the computational cost of Video LLMs, the compression process itself may introduce additional overhead. We evaluate the efficiency of SimpleComp on LLaVA-OneVision-7B, LLaVA-Video-7B, and InternVL3-8B at a 10\% token retention ratio, comparing it with the recent strong baseline AOT. All experiments are conducted on a single NVIDIA A800 GPU using the same video-prompt pairs. We report the theoretical floating-point operations (FLOPs) of the vision encoder and LLM prefilling, as well as time to first token (TTFT). TTFT includes visual encoding, multimodal projection, token compression, and first-token generation, but excludes model loading, video decoding, and preprocessing. TTFT is averaged over five measurement runs.

As shown in Table~\ref{tab:efficiency}, SimpleComp reduces prefill FLOPs by 77.2\%, 73.2\%, and 83.1\% on LLaVA-OneVision-7B, LLaVA-Video-7B, and InternVL3-8B, respectively, compared with the uncompressed models. Meanwhile, TTFT is reduced by 45.5\%, 37.7\%, and 56.0\%, respectively. Compared with AOT, SimpleComp further reduces TTFT by 26.2\%, 6.1\%, and 14.2\% across the three models. Unlike AOT, SimpleComp performs cross-frame feature-space clustering without attention-guided anchor selection or optimal transport matching. Moreover, SimpleComp retains 97.9\%, 94.8\%, and 96.1\% of the uncompressed models' average performance across five video understanding benchmarks, demonstrating its effectiveness in balancing computational efficiency and video understanding performance.

%% file: tabs/llava_ov.tex
\begin{table*}[t]
\caption{Comparison with state-of-the-art methods on LLaVA-OneVision-7B. ``Avg. Score'' and ``Avg. \%'' denote the average score across benchmarks and the relative performance to the unpruned model. FastVID and AOT are omitted at 1\% because their structural constraints require more tokens than the available budget. The best and second-best results are in \textbf{bold} and \underline{underlined}, respectively.}
\label{tab:llava_ov}
\centering
\resizebox{\textwidth}{!}{%
\begin{tabular}{l|c|c|ccccc|cc}
\toprule
\multirow{2}{*}{Method}
& \multirow{2}{*}{Source}
& \multirow{2}{*}{\shortstack{Before LLM\\Retained Ratio}}
& \multirow{2}{*}{\shortstack{VideoMME\\$\uparrow$}}
& \multirow{2}{*}{\shortstack{LongVideoBench\\$\uparrow$}}
& \multirow{2}{*}{\shortstack{MLVU\\$\uparrow$}}
& \multirow{2}{*}{\shortstack{MVBench\\$\uparrow$}}
& \multirow{2}{*}{\shortstack{EgoSchema\\$\uparrow$}}
& \multicolumn{2}{|c}{Avg. $\uparrow$} \\
& & & & & & & & Score & \% \\
\midrule
\rowcolor[RGB]{235,235,235}
LLaVA-OV-7B~\cite{li2025llava} & -- & 100\% & 58.6 & 56.4 & 63.0 & 58.3 & 60.4 & 59.3 & 100.0 \\
\midrule
FastVID~\cite{shen2025fastvid} & NeurIPS'25 & 15\% & \textbf{57.7} & \underline{56.2} & 61.0 & 56.0 & 57.4 & 57.7 & 97.2 \\
VisionZip~\cite{yang2025visionzip} & CVPR'25 & 15\% & 56.1 & 54.4 & 60.8 & 56.5 & 59.8 & 57.5 & 96.9 \\
EarlyTom~\cite{wang2026earlytom} & CVPR'26 & 15\% & 56.9 & 54.4 & \underline{61.2} & \underline{57.5} & \underline{60.2} & 58.0 & 97.8 \\
AOT~\cite{li2026token} & CVPR'26 & 15\% & 56.6 & 55.2 & 60.1 & \textbf{57.8} & \textbf{61.3} & \underline{58.2} & \underline{98.1} \\
\rowcolor[RGB]{216,229,223}
\textbf{SimpleComp} & -- & 15\% & \underline{57.3} & \textbf{56.9} & \textbf{61.8} & 57.4 & 60.1 & \textbf{58.7} & \textbf{98.9} \\
\midrule
FastVID~\cite{shen2025fastvid} & NeurIPS'25 & 10\% & \textbf{57.3} & \underline{56.3} & \underline{61.1} & 55.9 & 56.5 & \underline{57.4} & \underline{96.7} \\
VisionZip~\cite{yang2025visionzip} & CVPR'25 & 10\% & 53.4 & 49.3 & 58.9 & 53.5 & 58.0 & 54.6 & 92.0 \\
EarlyTom~\cite{wang2026earlytom} & CVPR'26 & 10\% & 55.8 & 52.4 & 59.8 & 56.5 & \underline{60.1} & 56.9 & 95.9 \\
AOT~\cite{li2026token} & CVPR'26 & 10\% & \underline{56.1} & 54.2 & 58.2 & \textbf{57.0} & \textbf{60.6} & 57.2 & 96.4 \\
\rowcolor[RGB]{216,229,223}
\textbf{SimpleComp} & -- & 10\% & 56.0 & \textbf{56.5} & \textbf{61.2} & \underline{56.8} & 59.9 & \textbf{58.1} & \textbf{97.9} \\
\midrule
FastVID~\cite{shen2025fastvid} & NeurIPS'25 & 5\% & \underline{53.8} & \underline{51.7} & \underline{57.6} & 53.8 & 56.9 & \underline{54.8} & \underline{92.3} \\
VisionZip~\cite{yang2025visionzip} & CVPR'25 & 5\% & 48.6 & 46.5 & 54.1 & 45.3 & 51.8 & 49.3 & 83.0 \\
EarlyTom~\cite{wang2026earlytom} & CVPR'26 & 5\% & 51.4 & 50.2 & 57.2 & 51.9 & 58.2 & 53.8 & 90.6 \\
AOT~\cite{li2026token} & CVPR'26 & 5\% & 52.3 & 50.5 & 56.8 & \underline{54.7} & \textbf{59.5} & \underline{54.8} & \underline{92.3} \\
\rowcolor[RGB]{216,229,223}
\textbf{SimpleComp} & -- & 5\% & \textbf{55.1} & \textbf{52.4} & \textbf{60.3} & \textbf{56.1} & \underline{59.1} & \textbf{56.6} & \textbf{95.4} \\
\midrule
VisionZip~\cite{yang2025visionzip} & CVPR'25 & 1\% & 44.4 & 44.7 & \underline{51.7} & 40.6 & 43.7 & 45.0 & 75.9 \\
EarlyTom~\cite{wang2026earlytom} & CVPR'26 & 1\% & \underline{45.9} & \underline{45.3} & 50.9 & \underline{43.3} & \underline{48.1} & \underline{46.7} & \underline{78.7} \\
\rowcolor[RGB]{216,229,223}
\textbf{SimpleComp} & -- & 1\% & \textbf{51.1} & \textbf{49.7} & \textbf{55.2} & \textbf{51.6} & \textbf{53.9} & \textbf{52.3} & \textbf{88.2} \\
\bottomrule
\end{tabular}%
}
\end{table*}

%% file: tabs/llava_video.tex
\begin{table*}[t]
\caption{Comparison with state-of-the-art methods on LLaVA-Video-7B. ``Avg. Score'' and ``Avg. \%'' denote the average score across benchmarks and the relative performance to the unpruned model. AOT is omitted at 1\% because its structural constraints require more tokens than the available budget. The best and second-best results are in \textbf{bold} and \underline{underlined}, respectively.}
\label{tab:llava_video}
\centering
\resizebox{\textwidth}{!}{%
\begin{tabular}{l|c|c|ccccc|cc}
\toprule
\multirow{2}{*}{Method}
& \multirow{2}{*}{Source}
& \multirow{2}{*}{\shortstack{Before LLM\\Retained Ratio}}
& \multirow{2}{*}{\shortstack{VideoMME\\$\uparrow$}}
& \multirow{2}{*}{\shortstack{LongVideoBench\\$\uparrow$}}
& \multirow{2}{*}{\shortstack{MLVU\\$\uparrow$}}
& \multirow{2}{*}{\shortstack{MVBench\\$\uparrow$}}
& \multirow{2}{*}{\shortstack{EgoSchema\\$\uparrow$}}
& \multicolumn{2}{|c}{Avg. $\uparrow$} \\
& & & & & & & & Score & \% \\
\midrule
\rowcolor[RGB]{235,235,235}
LLaVA-Video-7B~\cite{zhang2025llava} & -- & 100\% & 64.3 & 58.9 & 68.6 & 60.4 & 57.2 & 61.9 & 100.0 \\
\midrule
VisionZip~\cite{yang2025visionzip} & CVPR'25 & 15\% & 60.7 & 54.7 & \underline{64.6} & 56.7 & \underline{54.7} & 58.3 & 94.2 \\
EarlyTom~\cite{wang2026earlytom} & CVPR'26 & 15\% & \underline{61.3} & 53.9 & 62.0 & 55.8 & \underline{54.7} & 57.5 & 93.0 \\
AOT~\cite{li2026token} & CVPR'26 & 15\% & \textbf{62.0} & \underline{55.0} & 63.7 & \underline{57.8} & \textbf{55.2} & \underline{58.7} & \underline{94.9} \\
\rowcolor[RGB]{216,229,223}
\textbf{SimpleComp} & -- & 15\% & 61.1 & \textbf{57.7} & \textbf{64.7} & \textbf{59.2} & 53.2 & \textbf{59.2} & \textbf{95.6} \\
\midrule
VisionZip~\cite{yang2025visionzip} & CVPR'25 & 10\% & 59.0 & 53.9 & \underline{62.8} & 57.8 & \underline{53.1} & 57.3 & 92.6 \\
EarlyTom~\cite{wang2026earlytom} & CVPR'26 & 10\% & 56.0 & 50.9 & 57.0 & 56.8 & 47.0 & 53.5 & 86.5 \\
AOT~\cite{li2026token} & CVPR'26 & 10\% & \textbf{60.8} & \underline{54.1} & 62.5 & \underline{59.1} & \textbf{53.4} & \underline{58.0} & \underline{93.7} \\
\rowcolor[RGB]{216,229,223}
\textbf{SimpleComp} & -- & 10\% & \underline{60.7} & \textbf{57.3} & \textbf{64.0} & \textbf{59.2} & 52.0 & \textbf{58.6} & \textbf{94.8} \\
\midrule
VisionZip~\cite{yang2025visionzip} & CVPR'25 & 5\% & 53.6 & \underline{50.1} & 56.0 & 49.9 & \underline{44.9} & 50.9 & 82.3 \\
EarlyTom~\cite{wang2026earlytom} & CVPR'26 & 5\% & 53.7 & 47.9 & 55.3 & 48.1 & 44.7 & 49.9 & 80.7 \\
AOT~\cite{li2026token} & CVPR'26 & 5\% & \underline{56.8} & 49.4 & \underline{59.0} & \underline{52.6} & \textbf{50.1} & \underline{53.6} & \underline{86.6} \\
\rowcolor[RGB]{216,229,223}
\textbf{SimpleComp} & -- & 5\% & \textbf{59.2} & \textbf{55.0} & \textbf{62.2} & \textbf{58.0} & \textbf{50.1} & \textbf{56.9} & \textbf{92.0} \\
\midrule
VisionZip~\cite{yang2025visionzip} & CVPR'25 & 1\% & \underline{46.6} & \underline{45.0} & \underline{50.8} & \underline{43.4} & \underline{36.5} & \underline{44.5} & \underline{71.8} \\
EarlyTom~\cite{wang2026earlytom} & CVPR'26 & 1\% & 44.6 & 43.4 & 48.4 & 41.8 & 33.0 & 42.2 & 68.3 \\
\rowcolor[RGB]{216,229,223}
\textbf{SimpleComp} & -- & 1\% & \textbf{54.7} & \textbf{50.6} & \textbf{56.6} & \textbf{54.1} & \textbf{44.7} & \textbf{52.1} & \textbf{84.3} \\
\bottomrule
\end{tabular}%
}
\end{table*}

%% file: tabs/internvl3.tex
\begin{table*}[!ht]
\caption{Comparison with state-of-the-art methods on InternVL3-8B. ``Avg. Score'' and ``Avg. \%'' denote the average score across benchmarks and the relative performance to the unpruned model. AOT is omitted at 1\% because its structural constraints require more tokens than the available budget. The best and second-best results are in \textbf{bold} and \underline{underlined}, respectively.}
\label{tab:internvl3}
\centering
\resizebox{\textwidth}{!}{%
\begin{tabular}{l|c|c|ccccc|cc}
\toprule
\multirow{2}{*}{Method}
& \multirow{2}{*}{Source}
& \multirow{2}{*}{\shortstack{Before LLM\\Retained Ratio}}
& \multirow{2}{*}{\shortstack{VideoMME\\$\uparrow$}}
& \multirow{2}{*}{\shortstack{LongVideoBench\\$\uparrow$}}
& \multirow{2}{*}{\shortstack{MLVU\\$\uparrow$}}
& \multirow{2}{*}{\shortstack{MVBench\\$\uparrow$}}
& \multirow{2}{*}{\shortstack{EgoSchema\\$\uparrow$}}
& \multicolumn{2}{|c}{Avg. $\uparrow$} \\
& & & & & & & & Score & \% \\
\midrule
\rowcolor[RGB]{235,235,235}
InternVL3-8B~\cite{zhu2025internvl3} & -- & 100\% & 64.9 & 58.7 & 68.8 & 74.1 & 62.6 & 65.8 & 100.0 \\
\midrule
VisionZip~\cite{yang2025visionzip} & CVPR'25 & 15\% & \underline{63.3} & \textbf{57.4} & \underline{67.8} & \underline{72.1} & \underline{60.0} & \textbf{64.1} & \textbf{97.4} \\
EarlyTom~\cite{wang2026earlytom} & CVPR'26 & 15\% & 62.0 & \textbf{57.4} & 67.6 & \textbf{72.8} & \textbf{60.1} & \underline{64.0} & \underline{97.2} \\
AOT~\cite{li2026token} & CVPR'26 & 15\% & 60.1 & 54.4 & 63.9 & 71.8 & 57.4 & 61.5 & 93.5 \\
\rowcolor[RGB]{216,229,223}
\textbf{SimpleComp} & -- & 15\% & \textbf{63.7} & \underline{56.6} & \textbf{67.9} & 71.8 & 59.6 & 63.9 & 97.1 \\
\midrule
VisionZip~\cite{yang2025visionzip} & CVPR'25 & 10\% & \underline{62.3} & \underline{55.2} & \underline{66.1} & 68.2 & \textbf{58.6} & \underline{62.1} & \underline{94.3} \\
EarlyTom~\cite{wang2026earlytom} & CVPR'26 & 10\% & 60.3 & \textbf{55.7} & 63.9 & \underline{70.6} & 58.1 & 61.7 & 93.8 \\
AOT~\cite{li2026token} & CVPR'26 & 10\% & 59.7 & 54.0 & 62.7 & 69.2 & 55.0 & 60.1 & 91.3 \\
\rowcolor[RGB]{216,229,223}
\textbf{SimpleComp} & -- & 10\% & \textbf{62.9} & \textbf{55.7} & \textbf{68.2} & \textbf{71.4} & \underline{58.2} & \textbf{63.3} & \textbf{96.1} \\
\midrule
VisionZip~\cite{yang2025visionzip} & CVPR'25 & 5\% & \underline{59.3} & 52.0 & \underline{62.9} & 61.7 & \underline{55.8} & 58.3 & 88.6 \\
EarlyTom~\cite{wang2026earlytom} & CVPR'26 & 5\% & 57.2 & \underline{53.8} & 62.8 & \underline{66.1} & 55.5 & \underline{59.1} & \underline{89.8} \\
AOT~\cite{li2026token} & CVPR'26 & 5\% & 56.6 & 51.9 & 60.2 & 65.3 & 50.9 & 57.0 & 86.6 \\
\rowcolor[RGB]{216,229,223}
\textbf{SimpleComp} & -- & 5\% & \textbf{61.9} & \textbf{56.0} & \textbf{67.0} & \textbf{69.3} & \textbf{55.9} & \textbf{62.0} & \textbf{94.2} \\
\midrule
VisionZip~\cite{yang2025visionzip} & CVPR'25 & 1\% & 50.3 & 46.6 & 53.0 & 46.1 & 44.1 & 48.0 & 73.0 \\
EarlyTom~\cite{wang2026earlytom} & CVPR'26 & 1\% & \underline{53.0} & \underline{49.2} & \underline{55.3} & \underline{51.7} & \underline{44.3} & \underline{50.7} & \underline{77.0} \\
\rowcolor[RGB]{216,229,223}
\textbf{SimpleComp} & -- & 1\% & \textbf{57.6} & \textbf{53.6} & \textbf{61.2} & \textbf{62.4} & \textbf{49.6} & \textbf{56.9} & \textbf{86.4} \\
\bottomrule
\end{tabular}%
}
\end{table*}

%% file: tabs/diff_baselines.tex
\begin{table*}[t]
\caption{Comparison with simple baselines on LLaVA-OneVision-7B. ``Avg. Score'' and ``Avg. \%'' denote the average score across benchmarks and the relative performance to the unpruned model. The best and second-best results are in \textbf{bold} and \underline{underlined}, respectively.}
\label{tab:diff_baselines}
\centering
\resizebox{\textwidth}{!}{%
\begin{tabular}{l|c|ccccc|cc}
\toprule
\multirow{2}{*}{Method}
& \multirow{2}{*}{\shortstack{Before LLM\\Retained Ratio}}
& \multirow{2}{*}{\shortstack{VideoMME\\$\uparrow$}}
& \multirow{2}{*}{\shortstack{LongVideoBench\\$\uparrow$}}
& \multirow{2}{*}{\shortstack{MLVU\\$\uparrow$}}
& \multirow{2}{*}{\shortstack{MVBench\\$\uparrow$}}
& \multirow{2}{*}{\shortstack{EgoSchema\\$\uparrow$}}
& \multicolumn{2}{|c}{Avg. $\uparrow$} \\
& & & & & & & Score & \% \\
\midrule
\rowcolor[RGB]{235,235,235}
LLaVA-OV-7B~\cite{li2025llava} & 100\% & 58.6 & 56.4 & 63.0 & 58.3 & 60.4 & 59.3 & 100.0 \\
\midrule
Spatial Average Pooling
& 10\% & 54.1 & 54.2 & 58.7 & 53.3 & 57.7 & 55.6 & 93.7 \\
Spatiotemporal Average Pooling
& 10\% & 53.5 & 54.4 & 58.9 & 54.1 & 57.4 & 55.7 & 93.8 \\
Spatial Grid Sampling
& 10\% & \underline{55.0} & \underline{54.8} & \underline{59.2} & 53.9 & 58.5 & \underline{56.3} & \underline{94.9} \\
Spatiotemporal Grid Sampling
& 10\% & 54.8 & 53.1 & 58.0 & \underline{55.0} & \underline{58.8} & 55.9 & 94.3 \\
\rowcolor[RGB]{216,229,223}
\textbf{SimpleComp}
& 10\% & \textbf{56.0} & \textbf{56.5} & \textbf{61.2} & \textbf{56.8} & \textbf{59.9} & \textbf{58.1} & \textbf{97.9} \\
\midrule
Spatial Average Pooling
& 5\% & 51.4 & 49.7 & \underline{57.4} & 49.0 & 53.5 & 52.2 & 88.0 \\
Spatiotemporal Average Pooling
& 5\% & 50.4 & 49.4 & 55.1 & 51.4 & 54.7 & 52.2 & 88.0 \\
Spatial Grid Sampling
& 5\% & 51.7 & \underline{51.5} & 56.8 & 50.1 & 56.3 & 53.3 & 89.8 \\
Spatiotemporal Grid Sampling
& 5\% & \underline{52.6} & 50.9 & 55.4 & \underline{52.4} & \underline{56.4} & \underline{53.5} & \underline{90.2} \\
\rowcolor[RGB]{216,229,223}
\textbf{SimpleComp}
& 5\% & \textbf{55.1} & \textbf{52.4} & \textbf{60.3} & \textbf{56.1} & \textbf{59.1} & \textbf{56.6} & \textbf{95.4} \\
\midrule
Spatial Average Pooling
& 1\% & 45.8 & 45.4 & 51.7 & 43.0 & 44.7 & 46.1 & 77.7 \\
Spatiotemporal Average Pooling
& 1\% & 45.9 & 45.3 & 52.2 & 45.6 & 47.0 & 47.2 & 79.5 \\
Spatial Grid Sampling
& 1\% & 46.4 & \underline{45.9} & 52.8 & 42.2 & 47.5 & 47.0 & 79.1 \\
Spatiotemporal Grid Sampling
& 1\% & \underline{48.8} & 45.0 & \underline{53.1} & \underline{46.8} & \underline{50.8} & \underline{48.9} & \underline{82.4} \\
\rowcolor[RGB]{216,229,223}
\textbf{SimpleComp}
& 1\% & \textbf{51.1} & \textbf{49.7} & \textbf{55.2} & \textbf{51.6} & \textbf{53.9} & \textbf{52.3} & \textbf{88.1} \\
\bottomrule
\end{tabular}%
}
\end{table*}

%% file: tabs/ablations.tex
\begin{table}[t]
\caption{Ablation studies on LLaVA-OneVision-7B at a 10\% visual token retention ratio. The default configuration is in gray.}
\label{tab:ablation}
\centering
\resizebox{\columnwidth}{!}{%
\begin{tabular}{l|ccccc}
\toprule
\multirow{2}{*}{Setting}
& VideoMME & LVB & MLVU & MVBench & EgoSchema \\
& $\uparrow$ & $\uparrow$ & $\uparrow$ & $\uparrow$ & $\uparrow$ \\
\midrule
None
& 55.8 & 55.4 & 60.2 & 55.8 & 59.1 \\
\rowcolor[RGB]{235,235,235}
3D RoPE
& \textbf{56.0} & \textbf{56.5} & \textbf{61.2} & \textbf{56.8} & \textbf{59.9} \\
\midrule
Per-Frame
& 55.8 & 54.1 & 60.7 & 55.0 & 59.3 \\
\rowcolor[RGB]{235,235,235}
Cross-Frame
& \textbf{56.0} & \textbf{56.5} & \textbf{61.2} & \textbf{56.8} & \textbf{59.9} \\
\midrule
Nearest
& \textbf{56.4} & 55.7 & 61.0 & \textbf{57.0} & 59.7 \\
\rowcolor[RGB]{235,235,235}
Mean
& 56.0 & \textbf{56.5} & \textbf{61.2} & 56.8 & \textbf{59.9} \\
\bottomrule
\end{tabular}%
}
\end{table}

%% file: tabs/efficiency.tex
\begin{table}[t]
\centering
\caption{Efficiency analysis of SimpleComp across different Video LLMs at a 10\% visual token retention ratio.}
\label{tab:efficiency}
\resizebox{\columnwidth}{!}{%
\begin{tabular}{l|cc|cc}
\toprule
\multirow{2}{*}{Method}
& \multirow{2}{*}{\shortstack{Prefilling\\FLOPs (T) $\downarrow$}}
& \multirow{2}{*}{\shortstack{TTFT\\(ms) $\downarrow$}}
& \multicolumn{2}{c}{Avg. $\uparrow$} \\
& & & Score & \% \\
\midrule
LLaVA-OV-7B~\cite{li2025llava} & 125.6 & 943.3 & 59.3 & 100.0 \\
AOT~\cite{li2026token} & 30.8 & 696.5 & 57.2 & 96.4 \\
\rowcolor[RGB]{216,229,223}
\textbf{SimpleComp} & \textbf{28.7} & \textbf{514.1} & \textbf{58.1} & \textbf{97.9} \\
\midrule
LLaVA-Video-7B~\cite{zhang2025llava} & 260.9 & 1730.9 & 61.9 & 100.0 \\
AOT~\cite{li2026token} & 71.8 & 1148.7 & 58.0 & 93.7 \\
\rowcolor[RGB]{216,229,223}
\textbf{SimpleComp} & \textbf{70.0} & \textbf{1078.5} & \textbf{58.6} & \textbf{94.8} \\
\midrule
InternVL3-8B~\cite{zhu2025internvl3} & 642.5 & 2611.5 & 65.8 & 100.0 \\
AOT~\cite{li2026token} & 108.5 & 1339.7 & 60.1 & 91.3 \\
\rowcolor[RGB]{216,229,223}
\textbf{SimpleComp} & \textbf{108.3} & \textbf{1150.0} & \textbf{63.3} & \textbf{96.1} \\
\bottomrule
\end{tabular}%
}
\end{table}

%% file: secs/5_analysis.tex
\section{Why Does Feature-Space Clustering Work?}
\label{sec:analysis}

Despite its simple design, SimpleComp achieves strong performance under aggressive video token compression. To understand this behavior, we analyze how different compression methods preserve the original visual feature distribution in terms of local approximation fidelity and global coverage, and examine their relationship with downstream performance.

\noindent \textbf{Local Fidelity and Global Coverage.}
To evaluate feature-space preservation, we compare the original dense visual tokens with the compressed tokens produced by different methods on LLaVA-OneVision-7B at 10\%, 5\%, and 1\% token retention ratios. We consider two complementary metrics: (1) L2 Normalized Quantization Error (L2-NQE) measures the distance between each original token and its nearest compressed token, reflecting local approximation fidelity. (2) Cosine Coverage@0.90 measures the fraction of original tokens for which at least one compressed token achieves a cosine similarity above 0.90, reflecting global coverage of the original feature distribution. For each benchmark, we first deduplicate the videos and average each metric across all videos, then report the equally weighted average across VideoMME, LongVideoBench, MLVU, MVBench, and EgoSchema.

\input{tabs/analysis}

As shown in Table~\ref{tab:analysis}, SimpleComp achieves the lowest L2-NQE and highest cosine coverage at all three retention ratios. At 10\% retention, SimpleComp achieves an L2-NQE of 0.2888 and a cosine coverage of 56.84\%, reducing the quantization error by 28.0\% compared with AOT (0.4010) and improving coverage by 23.60 percentage points over FastVID (33.24\%). At 5\% retention, SimpleComp maintains an L2-NQE of 0.3444 and a cosine coverage of 43.09\%, compared with the best competing results of 0.4890 from AOT and 19.99\% from FastVID. Notably, SimpleComp at 5\% retention still achieves better feature-space preservation than all competing methods at 10\% retention. Even at an extremely low retention ratio of 1\%, SimpleComp maintains an L2-NQE of 0.4647 and a cosine coverage of 20.65\%, substantially outperforming VisionZip and EarlyTom. These results demonstrate that SimpleComp effectively preserves both local approximation fidelity and global coverage of the original visual feature distribution, even under aggressive compression.

\noindent \textbf{From Feature Preservation to Downstream Performance.}
The feature-space preservation results are consistent with the downstream performance trends in Table~\ref{tab:llava_ov}. At 10\%, 5\%, and 1\% retention, SimpleComp achieves the highest average downstream scores of 58.1, 56.6, and 52.3, respectively, while consistently maintaining the lowest L2-NQE and highest cosine coverage. In contrast, competing methods generally exhibit larger performance degradation as the token budget decreases. These observations suggest that better preservation of the original visual feature distribution, particularly its global coverage, is closely associated with stronger downstream video understanding performance.

This relationship becomes more evident under increasingly aggressive compression. When the retention ratio decreases from 10\% to 5\%, SimpleComp exhibits an L2-NQE increase of only 0.0556 and retains 75.8\% of its original cosine coverage, compared with L2-NQE increases of 0.088--0.148 and coverage retention of 42.8\%--63.0\% for competing methods. Correspondingly, its average downstream score decreases by only 1.5 points, while competing methods experience drops of 2.4--5.3 points. Furthermore, at 1\% retention, SimpleComp achieves an average score of 52.3, substantially outperforming the best applicable competing method, EarlyTom (46.7). These consistent trends further highlight the importance of preserving the original feature-space structure for maintaining downstream performance under highly constrained token budgets.

%% file: tabs/analysis.tex
\begin{table}[t]
\centering
\caption{Comparison of local fidelity and global coverage of the original visual feature space across different visual token compression methods on LLaVA-OneVision-7B. Results are averaged across five benchmarks: VideoMME, LongVideoBench, MLVU, MVBench, and EgoSchema.}
\label{tab:analysis}
\resizebox{\columnwidth}{!}{%
\begin{tabular}{l|c|cc}
\toprule
\multirow{2}{*}{Method}
& \multirow{2}{*}{\shortstack{Before LLM\\Retention}}
& \multirow{2}{*}{\shortstack{L2-NQE\\$\downarrow$}}
& \multirow{2}{*}{\shortstack{Cosine Coverage\\@0.90 (\%) $\uparrow$}} \\
& & & \\
\midrule
VisionZip~\cite{yang2025visionzip} & 10\% & 0.5578 & 18.28 \\
FastVID~\cite{shen2025fastvid} & 10\% & 0.4168 & 33.24 \\
EarlyTom~\cite{wang2026earlytom} & 10\% & 0.4158 & 27.18 \\
AOT~\cite{li2026token} & 10\% & 0.4010 & 22.51 \\
\rowcolor[RGB]{216,229,223}
\textbf{SimpleComp} & 10\% & \textbf{0.2888} & \textbf{56.84} \\
\midrule
VisionZip~\cite{yang2025visionzip} & 5\% & 0.6907 & 7.82 \\
FastVID~\cite{shen2025fastvid} & 5\% & 0.5643 & 19.99 \\
EarlyTom~\cite{wang2026earlytom} & 5\% & 0.5620 & 15.39 \\
AOT~\cite{li2026token} & 5\% & 0.4890 & 14.18 \\
\rowcolor[RGB]{216,229,223}
\textbf{SimpleComp} & 5\% & \textbf{0.3444} & \textbf{43.09} \\
\midrule
VisionZip~\cite{yang2025visionzip} & 1\% & 1.0201 & 2.49 \\
EarlyTom~\cite{wang2026earlytom} & 1\% & 1.0670 & 3.77 \\
\rowcolor[RGB]{216,229,223}
\textbf{SimpleComp} & 1\% & \textbf{0.4647} & \textbf{20.65} \\
\bottomrule
\end{tabular}%
}
\end{table}

%% file: secs/6_conclusion.tex
\section{Conclusion}
\label{sec:conclusion}
We present SimpleComp, a simple and training-free baseline for video token compression based on position-aware cross-frame clustering in the visual feature space. Across three representative Video LLMs and five video understanding benchmarks, SimpleComp outperforms recent compression methods at most token retention ratios, with particularly strong robustness under aggressive compression. To understand this behavior, we analyzed the feature space preserved by different compression methods and found that stronger downstream performance is consistently associated with better preservation of the original visual feature distribution, in terms of both local approximation fidelity and global coverage. These findings suggest that preserving the structure of the visual feature space is an important consideration for video token compression, particularly when the available token budget is highly constrained.

%% file: secs/7_limitation.tex
\section{Limitation}
\label{sec:limitation}

\noindent \textbf{Scope as a Strong Baseline Study.}
This work is positioned as a strong baseline study rather than a new video token compression algorithm. Although SimpleComp achieves strong performance across a wide range of settings, more sophisticated compression mechanisms may still be beneficial in certain scenarios. Our current study does not fully characterize when such designs provide additional advantages. Future work could explore adaptive compression strategies that explicitly account for feature-space coverage and local fidelity while balancing efficiency, generality, and computational cost.

%% file: secs/X_suppl.tex
\clearpage
\hypersetup{pageanchor=false}
\setcounter{page}{1}
\maketitlesupplementary

\setcounter{section}{0}
\renewcommand{\thesection}{\Alph{section}}
\renewcommand{\theHsection}{supp.\Alph{section}}

\section*{Overview}
This supplementary material provides additional evaluations, experimental analyses, implementation details, and qualitative visualizations to complement the main paper. The contents are organized as follows:
\begin{itemize}
\item \textbf{Section~\ref{sec:evaluation_benchmarks}} introduces the five evaluation benchmarks used in our experiments, including VideoMME, LongVideoBench, MLVU, MVBench, and EgoSchema. We summarize their key characteristics, covering diverse video durations, long-context comprehension, temporal reasoning, and egocentric understanding.
\item \textbf{Section~\ref{sec:more_experimental_results}} presents additional experimental results to further evaluate the effectiveness and generalizability of SimpleComp. We provide extended comparisons with existing visual token compression methods and simple compression strategies, evaluate its performance on a smaller Video LLM, and analyze its computational efficiency in terms of clustering overhead, compression time, and LLM prefill time.
\item \textbf{Section~\ref{sec:implementation-details}} provides additional implementation details to facilitate reproducibility. We describe the reproduction settings and backbone-specific adaptations of the compared methods, together with the visual token construction procedures. We further detail the position-aware feature construction and global cross-frame clustering of SimpleComp.
\item \textbf{Section~\ref{sec:more_visualizations}} presents additional qualitative visualizations of SimpleComp, including cross-frame clustering results across diverse videos, clustering assignments under different token retention ratios, and dense-feature coverage maps. These visualizations further illustrate the clustering behavior and its effectiveness in preserving visual feature information under constrained token budgets.
\end{itemize}

\section{Evaluation Benchmarks}
\label{sec:evaluation_benchmarks}
We evaluate SimpleComp on five widely used video understanding benchmarks covering temporal reasoning, egocentric understanding, and long-context video comprehension. These benchmarks encompass diverse video durations, scenarios, and tasks, enabling a comprehensive evaluation of visual token compression.

\noindent \textbf{VideoMME.}
VideoMME~\cite{fu2025video} evaluates video understanding across diverse content and temporal scales. It contains 900 videos spanning 6 major domains and 30 fine-grained categories, with durations ranging from 11 seconds to 1 hour. The benchmark includes 2,700 human-annotated multiple-choice questions assessing video perception, comprehension, and reasoning across short, medium, and long videos.

\noindent \textbf{LongVideoBench.}
LongVideoBench~\cite{wu2024longvideobench} evaluates long-context video-language understanding through 3,763 videos ranging from 8 seconds to 1 hour and 6,678 human-annotated multiple-choice questions. It includes 17 question categories based on referring reasoning, requiring models to identify relevant visual and textual evidence and reason over extended video sequences.

\noindent \textbf{MLVU.}
MLVU~\cite{zhou2025mlvu} evaluates multi-task long-video understanding across diverse video genres and durations ranging from 3 minutes to 2 hours. It covers nine evaluation tasks designed to assess holistic understanding, fine-grained detail recognition, and reasoning over long temporal contexts. These tasks provide a comprehensive assessment of models' ability to extract and integrate information from extended video sequences.

\noindent \textbf{MVBench.}
MVBench~\cite{li2024mvbench} evaluates temporal understanding through 20 video-centric tasks constructed by extending static image tasks to dynamic scenarios. These tasks cover action recognition, temporal localization, object interactions, and event reasoning, requiring models to capture temporal changes and relationships that cannot be inferred from individual frames.

\noindent \textbf{EgoSchema.}
EgoSchema~\cite{mangalam2023egoschema} evaluates long-horizon reasoning in egocentric videos. Derived from Ego4D, it contains over 5,000 human-annotated five-option multiple-choice questions spanning more than 250 hours of video data. Each question is associated with a three-minute video clip and requires models to understand extended human activities and reason about temporal relationships across events.

\input{tabs/sup_llava_ov}
\input{tabs/sup_llava_video}
\input{tabs/sup_internvl3}

\input{tabs/sup_diff_baselines}
\input{tabs/sup_internvl3_2b}

\input{tabs/sup_runtime}

\section{Additional Experimental Results}
\label{sec:more_experimental_results}
We provide additional experimental results to further evaluate the effectiveness, generalizability, and computational efficiency of SimpleComp. Specifically, we compare SimpleComp with recent compression methods at 25\% and 20\% token retention ratios, and with simple token compression strategies at 25\%, 20\%, and 15\% retention. We also extend our evaluation to the smaller InternVL3-2B model across six retention ratios. We further provide a detailed efficiency analysis on LLaVA-OneVision-7B and LLaVA-Video-7B, reporting runtime breakdowns, end-to-end latency, throughput, and peak GPU memory under different token budgets.

\noindent \textbf{Comparisons with Existing Compression Methods.}
Tables~\ref{tab:sup_llava_ov}, \ref{tab:sup_llava_video}, and \ref{tab:sup_internvl3} report additional comparisons with recent visual token compression methods at 25\% and 20\% retention. At 25\% retention, SimpleComp achieves the highest average scores of 59.6 and 60.4 on LLaVA-OneVision-7B and LLaVA-Video-7B, respectively, while obtaining a competitive score of 64.7 on InternVL3-8B. At 20\% retention, SimpleComp remains competitive across all three backbones, achieving average scores of 58.7, 59.9, and 64.3, respectively. These results further demonstrate the effectiveness of SimpleComp across different Video LLM backbones and token budgets.

\noindent \textbf{Comparisons with Simple Compression Strategies.}
We further compare SimpleComp with simple token compression strategies on LLaVA-OneVision-7B at 25\%, 20\%, and 15\% retention. As shown in Table~\ref{tab:sup_diff_baselines}, SimpleComp achieves the highest average scores of 59.6, 58.7, and 58.7, outperforming the strongest competing strategies by 1.4, 0.9, and 1.3 points, respectively. These results highlight the effectiveness of feature-space clustering across different token budgets.

\noindent \textbf{Performance on a Smaller Video LLM.}
To further evaluate the effectiveness of SimpleComp on smaller Video LLMs, we conduct additional experiments on InternVL3-2B across six visual token retention ratios (1\%, 5\%, 10\%, 15\%, 20\%, and 25\%). As shown in Table~\ref{tab:internvl3_2b}, SimpleComp consistently achieves the highest average scores across all retention ratios, demonstrating its effectiveness on a smaller backbone. Notably, its advantages become more pronounced under aggressive compression. At just 1\% retention, SimpleComp outperforms the strongest competing baseline by 4.3 points, retaining 86.8\% of the original model's performance. These results further demonstrate the robustness and generalizability of SimpleComp across different model scales, particularly under extremely limited visual token budgets.

\noindent \textbf{Detailed Efficiency Analysis.}
To further evaluate the computational overhead of cross-frame clustering, we compare SimpleComp with AOT under different token retention budgets. AOT is excluded from the 1\% retention setting because its structural constraints require more tokens than the available budget. Accordingly, we evaluate both methods at retention ratios of 5\%, 10\%, 15\%, 20\%, and 25\%, with the corresponding uncompressed models serving as references. Experiments are conducted on LLaVA-OneVision-7B and LLaVA-Video-7B, using their native configurations of 32 and 64 sampled frames, respectively. All measurements are performed on a single NVIDIA A800 80GB GPU with a batch size of 1 and BF16 precision, using the same video--prompt pair. Each configuration undergoes two warm-up runs followed by ten measured runs. We adopt greedy decoding with a fixed generation length of 256 tokens and disable EOS-based early stopping.

Tables~\ref{tab:sup_runtime_llava_ov} and \ref{tab:sup_runtime_llava_video} report the clustering core time, total compression time, and LLM prefill time on the two backbones, respectively. All runtimes are measured in milliseconds and averaged over ten runs. The clustering core includes feature normalization, prototype initialization, cluster assignment and update, and original-feature averaging, while total compression time additionally accounts for positional encoding, output ordering, and visual sequence preparation. On LLaVA-OneVision-7B, SimpleComp consistently achieves lower compression overhead than AOT across all evaluated retention ratios, reducing compression time by 77.4\% at 5\% retention (173.0\,ms to 39.1\,ms) and from 251.8\,ms to 101.4\,ms at 15\% retention. On LLaVA-Video-7B, SimpleComp also reduces compression time at retention ratios of 5\%--20\%, although it incurs slightly higher overhead at 25\% retention (461.1\,ms vs. 448.7\,ms). Regarding LLM prefill, SimpleComp substantially reduces processing time compared with the uncompressed models, particularly at aggressive compression ratios. At 5\% retention, prefill time decreases from 559.0\,ms to 46.8\,ms on LLaVA-OneVision-7B and from 1076.4\,ms to 123.7\,ms on LLaVA-Video-7B, demonstrating the effectiveness of visual token compression in reducing the computational burden of LLM prefill.

\section{Additional Implementation Details}
\label{sec:implementation-details}

\subsection{Reproduction Details of Compared Methods}
\label{sec:baseline-implementation}

We provide reproduction details for four representative compression baselines: VisionZip~\cite{yang2025visionzip}, FastVID~\cite{shen2025fastvid}, EarlyTom~\cite{wang2026earlytom}, and AOT~\cite{li2026token}. For LLaVA-OneVision-7B and LLaVA-Video-7B, we adopt the results reported in the original papers wherever available. We evaluate the remaining unreported settings, including additional token retention ratios and benchmark configurations, using the corresponding officially released implementations. We preserve each method's original compression procedures and follow its released hyperparameter configurations. For additional retention ratios, we adjust the native budget-related settings while keeping the remaining hyperparameters unchanged.

For InternVL3, we adapt the released implementations of the evaluated baselines to the corresponding backbone. These adaptations address model-specific interfaces, feature dimensions, visual-token layouts, and token insertion into the language model. We preserve the core token selection, matching, merging, and aggregation operations, and follow the original hyperparameter settings without additional backbone-specific tuning. The reported InternVL3 results are obtained using these adapted implementations.

For experiments reproduced using released implementations, we use consistent model checkpoints, frame sampling, text prompts, generation configurations, and benchmark scorers on the same backbone, while preserving method-specific processing required by the original implementations. Unless otherwise specified, token retention is measured by the number of visual content tokens immediately before LLM input. Model-specific structural tokens, including video newline and spatial-grid newline tokens, are excluded from the visual content-token budget.

All compression methods operate on the projected visual content tokens produced by their respective backbones. The vision encoder, multimodal projector, and native spatial pooling operations remain unchanged. Compression is applied after these operations and before the visual tokens enter the LLM. Model-specific structural tokens are retained without compression.

For LLaVA-OneVision-7B, we uniformly sample 32 frames. After native stride-2 bilinear spatial pooling, each frame produces a $14\times14$ visual-token grid, yielding
\begin{equation}
N = 32\times14\times14 = 6{,}272
\end{equation}
visual content tokens.

For LLaVA-Video-7B, we uniformly sample 64 frames following the fixed-64-frame evaluation protocol. Each frame produces a $13\times13$ content-token grid after native average pooling, yielding
\begin{equation}
N = 64\times13\times13 = 10{,}816
\end{equation}
visual content tokens. The original spatial-grid newline tokens remain unchanged and are excluded from the token budget.

For InternVL3-8B, we uniformly sample 32 frames and follow its official dynamic image-tiling procedure. Each tile produces a $16\times16$ projected token grid. Since the number of tiles varies across videos, the dense visual-token count is sample-dependent. All visual tokens from a video share a single compression budget.

The method-specific settings and necessary adaptations are detailed below.
\begin{itemize}
\item \textbf{VisionZip (CVPR 2025).}
Following VisionZip~\cite{yang2025visionzip}, we extract token importance scores and key features from the penultimate vision encoder layer. Attention-based selection identifies dominant tokens, while the remaining tokens are assigned to selected representatives to form contextual tokens. We preserve the original allocation between dominant and contextual tokens. For inputs containing multiple frames or image tiles, we distribute the video-level budget using integer largest-remainder allocation, maintaining non-empty dominant and contextual branches whenever the budget permits.
\item \textbf{FastVID (NeurIPS 2025).}
We reproduce FastVID~\cite{shen2025fastvid} using its released dynamic segmentation, spatiotemporal pruning, and dynamic token merging components. We set $c=8$ and $\tau=0.9$ for dynamic segmentation, $d=0.4$ for spatiotemporal pruning, and $p=4$ and $\beta=0.6$ for dynamic token merging. We retain the original frame-wise construction and merging procedures for salient and contextual tokens. At extremely low retention ratios, we report only configurations that produce valid, non-empty token representations, without additional truncation or padding.
\item \textbf{EarlyTom (CVPR 2026).}
We use the released EarlyTom~\cite{wang2026earlytom} implementation with its task-specific configurations. The exponential moving average coefficient is set to $0.9$, and the temporal matching range is $M=6$. The pruning threshold and layer schedule follow the corresponding benchmark and retention-ratio settings. We retain the internal token reduction at LLM layer 18 with a retention ratio of $0.5$. For vision encoders with different depths, pruning locations are mapped to their corresponding relative depths, preserving the operation order and pruning rules.
\item \textbf{AOT (CVPR 2026).}
We reproduce AOT~\cite{li2026token} using its original intra-frame and inter-frame optimal transport operations. The Sinkhorn regularization coefficient is set to $0.1$, with a maximum of 100 iterations, and both transport-cost scales are set to $1.0$. We preserve the backbone-specific keep ratios and anchor construction. Since AOT supports a discrete set of valid anchor configurations, its actual retention ratio may differ from the target. We select the valid configuration closest to the target ratio and report its token count without additional truncation or padding.
\end{itemize}

\subsection{Implementation Details of SimpleComp}
\label{sec:SimpleComp-implementation}

\noindent \textbf{Position-Aware Feature Construction.}
Let
\begin{equation}
X = \{x_i\}_{i=1}^{N},
\qquad
x_i \in \mathbb{R}^{d},
\end{equation}
denote the original projected visual content tokens. Each token is associated with a spatiotemporal position
\begin{equation}
p_i = (f_i,h_i,w_i),
\end{equation}
where $f_i$ denotes the frame index and $(h_i,w_i)$ specifies its spatial grid position.

We divide the feature channels into temporal, vertical, and horizontal groups and apply a 3D rotary transformation to encode spatiotemporal positions:
\begin{equation}
\widetilde{x}_i = R(p_i)x_i,
\end{equation}
where $R(p_i)$ is determined by $(f_i,h_i,w_i)$. We use a rotary frequency base of $\theta=10{,}000$. The transformed features are then L2-normalized:
\begin{equation}
z_i =
\frac{\widetilde{x}_i}
{\max\left(\lVert\widetilde{x}_i\rVert_2,\epsilon\right)}.
\end{equation}
The position-aware features $\{z_i\}_{i=1}^{N}$ are used for clustering, while the original features $\{x_i\}_{i=1}^{N}$ are retained for cluster-wise aggregation.

\noindent \textbf{Global Cross-Frame Clustering.}
Given a target retention ratio $r$, the number of clusters is set to
\begin{equation}
B =
\min\left(
N,
\max\left(
1,
\operatorname{round}(rN)
\right)
\right)
\end{equation}
visual groups. The resulting budget $B$ determines the number of aggregated visual tokens passed to the LLM.

Visual tokens from all sampled frames are jointly partitioned into $B$ clusters:
\begin{equation}
\mathcal{G}
=
\{\mathcal{C}_1,\ldots,\mathcal{C}_B\}.
\end{equation}
We optimize the following token-to-prototype cosine-distance objective:
\begin{equation}
\mathcal{L}_{\mathrm{cluster}}
=
\sum_{i=1}^{N}
\left(
1 -
\frac{
z_i^{\top}\mu_{a_i}
}{
\lVert z_i\rVert_2
\lVert \mu_{a_i}\rVert_2
}
\right),
\end{equation}
where $\mu_k$ denotes the prototype of cluster $\mathcal{C}_k$ and $a_i$ denotes the cluster assignment of token $i$.

For fixed prototypes, each token is assigned to the cluster with the highest cosine similarity:
\begin{equation}
a_i
=
\arg\max_{k\in\{1,\ldots,B\}}
\frac{
z_i^{\top}\mu_k
}{
\lVert z_i\rVert_2
\lVert \mu_k\rVert_2
}.
\end{equation}
For fixed assignments, each prototype is updated using the normalized mean of its position-aware features:
\begin{equation}
\mu_k
\leftarrow
\frac{
\sum_{i\in\mathcal{C}_k} z_i
}{
\left\lVert
\sum_{i\in\mathcal{C}_k} z_i
\right\rVert_2
}.
\end{equation}

The resulting assignments are used to aggregate the original projected features within each cluster, producing $B$ visual tokens for the LLM.

\section{Additional Visualizations}
\label{sec:more_visualizations}

We provide additional visualizations of clustering assignments and dense-feature coverage, illustrating how SimpleComp groups visual tokens across frames and preserves original visual features under different token budgets.

\subsection{Clustering Visualizations}

\begin{figure*}[t]
\centering
\includegraphics[width=\linewidth]{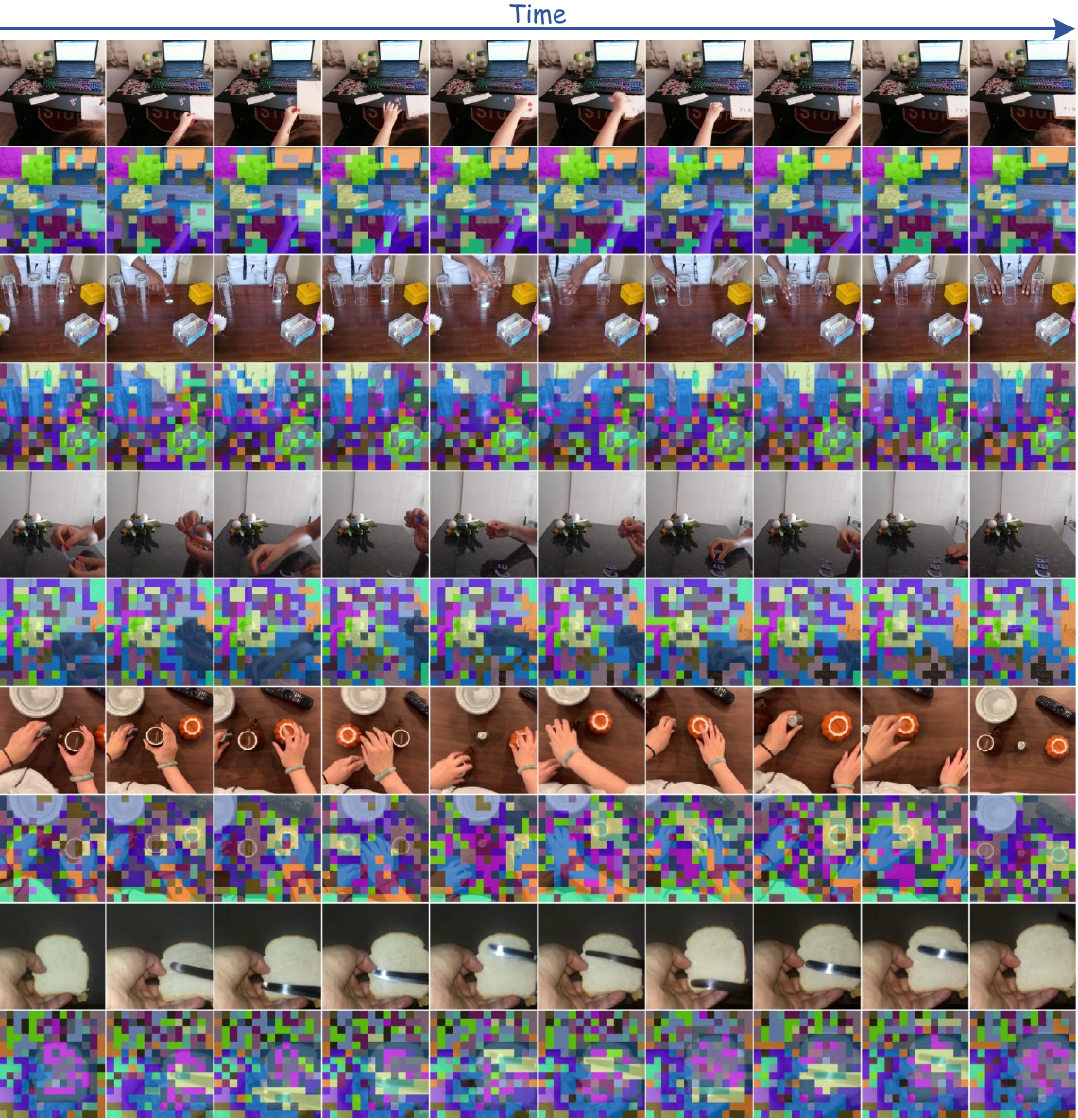}
\caption{Cross-frame clustering visualizations generated by SimpleComp at 5\% visual token retention. Each example shows video frames and their corresponding cluster assignments over time, where tokens with the same color belong to the same cluster.}
\label{fig:sup_fig1}
\end{figure*}

Figure~\ref{fig:sup_fig1} presents cross-frame clustering visualizations for five videos with diverse visual content and temporal dynamics. Each example shows video frames and their corresponding clustering results, where tokens assigned to the same cluster share the same color. SimpleComp effectively groups visually similar regions across frames while distinguishing different visual content, producing coherent and consistent clustering structures. These results demonstrate its ability to capture cross-frame feature similarities and exploit temporal redundancy.

\begin{figure*}[t]
\centering
\includegraphics[width=\linewidth]{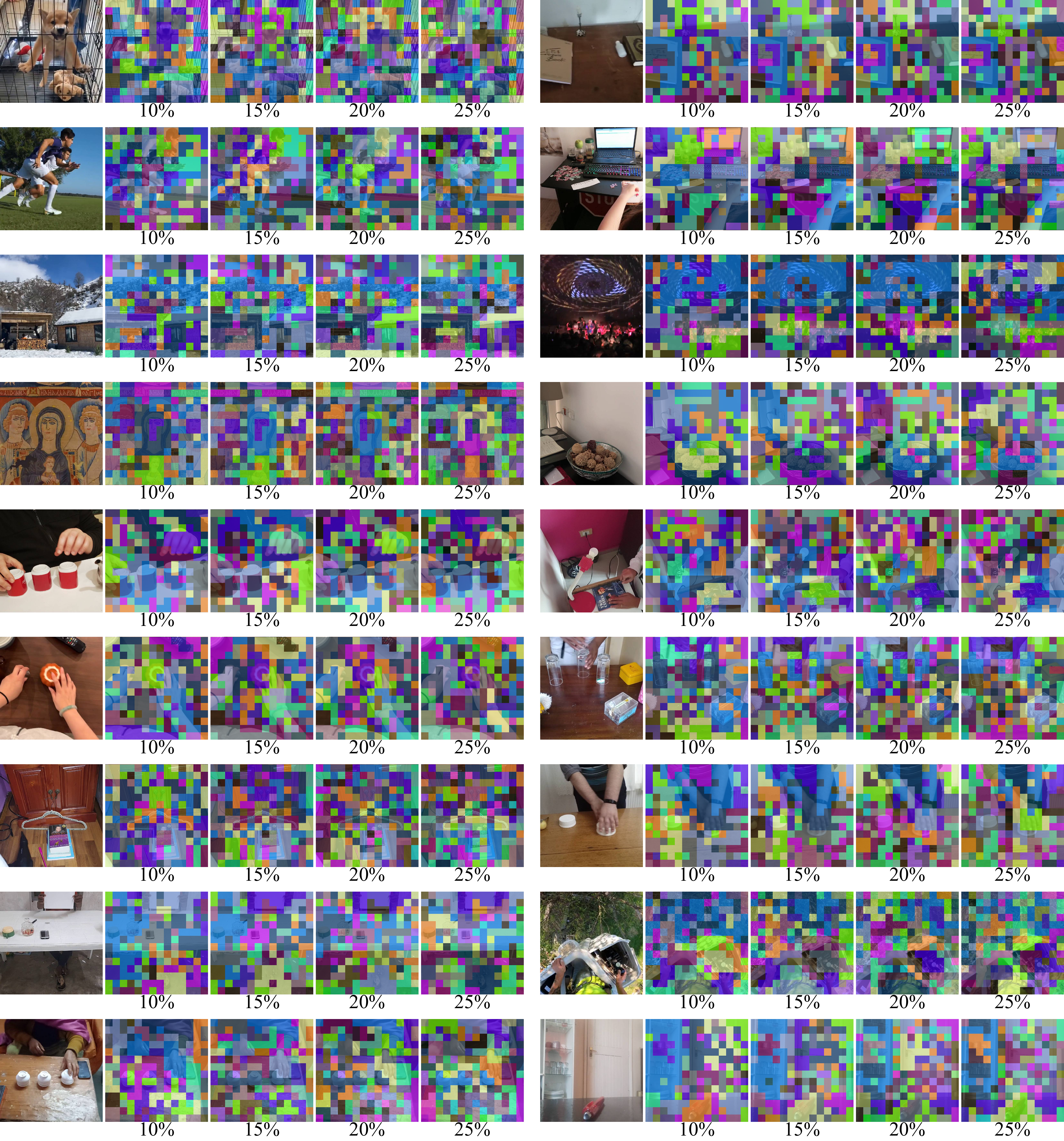}
\caption{Clustering visualizations of individual video frames generated by SimpleComp at different visual token retention ratios (10\%, 15\%, 20\%, and 25\%). Tokens with the same color belong to the same cluster within each visualization.}
\label{fig:sup_fig2}
\end{figure*}

Figure~\ref{fig:sup_fig2} further illustrates the clustering results for individual video frames at different visual token retention ratios (10\%, 15\%, 20\%, and 25\%). As the retention ratio increases, more clusters enable finer-grained visual grouping. SimpleComp maintains coherent clustering structures across different compression levels while preserving the major visual regions.

\subsection{Dense-Feature Coverage}

\begin{figure*}[t]
\centering
\includegraphics[width=0.6\linewidth]{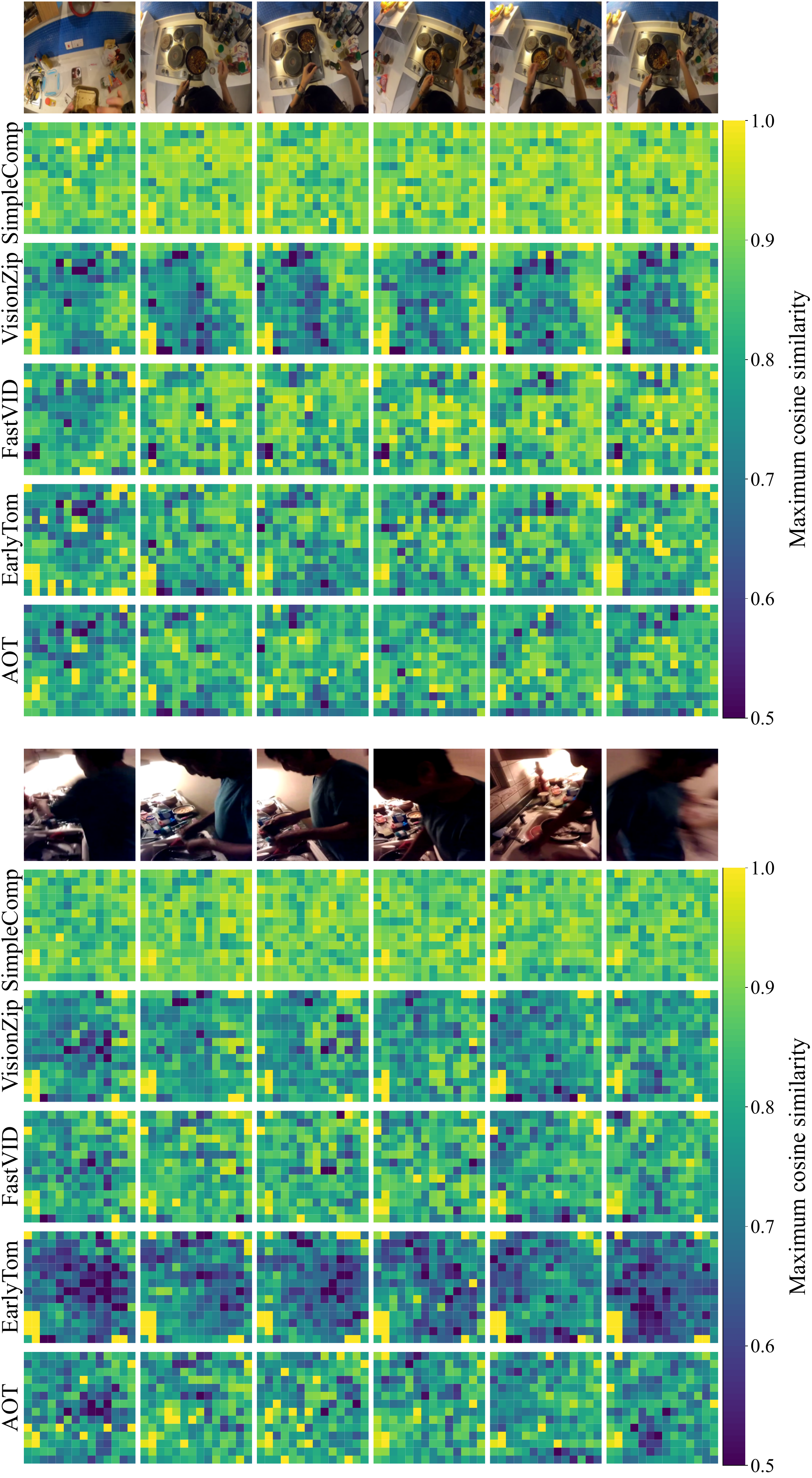}
\caption{Dense-feature coverage visualizations at 5\% visual token retention. Each heatmap shows the maximum cosine similarity between original dense visual tokens and compressed tokens produced by the corresponding method. Brighter colors indicate higher feature-space similarity.}
\label{fig:sup_cover_map}
\end{figure*}

Figure~\ref{fig:sup_cover_map} visualizes the coverage of original dense visual features at a 5\% visual token retention ratio. For each original token $x_i$, we calculate its maximum cosine similarity to the compressed tokens:
\begin{equation}
s_i =
\max_j
\frac{x_i^\top y_j}
{\lVert x_i\rVert_2\lVert y_j\rVert_2},
\end{equation}
where $y_j$ denotes a compressed token. The resulting similarity scores are mapped back to the original spatial token grids.

As shown in Figure~\ref{fig:sup_cover_map}, SimpleComp achieves consistently higher and more uniform feature similarity across different frames in both videos. In contrast, competing methods exhibit larger regions with relatively low similarity, indicating less effective coverage of the original dense visual features. These observations are consistent with the lower L2-NQE and higher Cosine Coverage@0.90 achieved by SimpleComp, further demonstrating its effectiveness in preserving visual feature information under constrained token budgets.

%% file: tabs/sup_llava_ov.tex
\begin{table*}[t]
\caption{Comparison with state-of-the-art methods on LLaVA-OneVision-7B. ``Avg. Score'' and ``Avg. \%'' denote the average score across benchmarks and the relative performance to the unpruned model. The best and second-best results are in \textbf{bold} and \underline{underlined}, respectively.}
\label{tab:sup_llava_ov}
\centering
\resizebox{\textwidth}{!}{%
\begin{tabular}{l|c|c|ccccc|cc}
\toprule
\multirow{2}{*}{Method}
& \multirow{2}{*}{Source}
& \multirow{2}{*}{\shortstack{Before LLM\\Retained Ratio}}
& \multirow{2}{*}{\shortstack{VideoMME\\$\uparrow$}}
& \multirow{2}{*}{\shortstack{LongVideoBench\\$\uparrow$}}
& \multirow{2}{*}{\shortstack{MLVU\\$\uparrow$}}
& \multirow{2}{*}{\shortstack{MVBench\\$\uparrow$}}
& \multirow{2}{*}{\shortstack{EgoSchema\\$\uparrow$}}
& \multicolumn{2}{|c}{Avg. $\uparrow$} \\
& & & & & & & & Score & \% \\
\midrule
\rowcolor[RGB]{235,235,235}
LLaVA-OV-7B~\cite{li2025llava} & -- & 100\% & 58.6 & 56.4 & 63.0 & 58.3 & 60.4 & 59.3 & 100.0 \\
\midrule
FastVID~\cite{shen2025fastvid} & NeurIPS'25 & 25\% & 58.0 & 56.3 & 61.7 & 56.5 & 58.2 & 58.1 & 98.0 \\
VisionZip~\cite{yang2025visionzip} & CVPR'25 & 25\% & 58.2 & \underline{56.5} & \underline{63.1} & 57.9 & 60.3 & \underline{59.2} & \underline{99.8} \\
EarlyTom~\cite{wang2026earlytom} & CVPR'26 & 25\% & \textbf{58.5} & 56.3 & 62.6 & 57.4 & \underline{60.5} & 59.1 & 99.5 \\
AOT~\cite{li2026token} & CVPR'26 & 25\% & 57.5 & 56.3 & 61.3 & \textbf{58.7} & \textbf{61.3} & 59.0 & 99.5 \\
\rowcolor[RGB]{216,229,223}
\textbf{SimpleComp} & -- & 25\% & \underline{58.4} & \textbf{57.6} & \textbf{63.8} & \underline{58.1} & 60.4 & \textbf{59.6} & \textbf{100.5} \\
\midrule
FastVID~\cite{shen2025fastvid} & NeurIPS'25 & 20\% & \underline{57.9} & \textbf{57.1} & 61.2 & 56.3 & 57.9 & 58.0 & 97.8 \\
VisionZip~\cite{yang2025visionzip} & CVPR'25 & 20\% & \underline{57.9} & 55.2 & 62.4 & 57.7 & 59.8 & 58.6 & 98.7 \\
EarlyTom~\cite{wang2026earlytom} & CVPR'26 & 20\% & \textbf{58.0} & 55.6 & \textbf{62.7} & \underline{57.8} & \underline{60.6} & \textbf{58.9} & \textbf{99.3} \\
AOT~\cite{li2026token} & CVPR'26 & 20\% & 57.2 & 56.2 & 60.9 & \textbf{58.1} & \textbf{61.3} & \underline{58.7} & \underline{98.9} \\
\rowcolor[RGB]{216,229,223}
\textbf{SimpleComp} & -- & 20\% & 57.6 & \underline{56.6} & \underline{62.5} & 57.0 & 59.8 & \underline{58.7} & \underline{98.9} \\
\bottomrule
\end{tabular}%
}
\end{table*}

%% file: tabs/sup_llava_video.tex
\begin{table*}[t]
\caption{Comparison with state-of-the-art methods on LLaVA-Video-7B. ``Avg. Score'' and ``Avg. \%'' denote the average score across benchmarks and the relative performance to the unpruned model. The best and second-best results are in \textbf{bold} and \underline{underlined}, respectively.}
\label{tab:sup_llava_video}
\centering
\resizebox{\textwidth}{!}{%
\begin{tabular}{l|c|c|ccccc|cc}
\toprule
\multirow{2}{*}{Method}
& \multirow{2}{*}{Source}
& \multirow{2}{*}{\shortstack{Before LLM\\Retained Ratio}}
& \multirow{2}{*}{\shortstack{VideoMME\\$\uparrow$}}
& \multirow{2}{*}{\shortstack{LongVideoBench\\$\uparrow$}}
& \multirow{2}{*}{\shortstack{MLVU\\$\uparrow$}}
& \multirow{2}{*}{\shortstack{MVBench\\$\uparrow$}}
& \multirow{2}{*}{\shortstack{EgoSchema\\$\uparrow$}}
& \multicolumn{2}{|c}{Avg. $\uparrow$} \\
& & & & & & & & Score & \% \\
\midrule
\rowcolor[RGB]{235,235,235}
LLaVA-Video-7B~\cite{zhang2025llava} & -- & 100\% & 64.3 & 58.9 & 68.6 & 60.4 & 57.2 & 61.9 & 100.0 \\
\midrule
VisionZip~\cite{yang2025visionzip} & CVPR'25 & 25\% & 60.7 & 54.7 & \underline{65.9} & 56.7 & 54.7 & 58.5 & 94.6 \\
EarlyTom~\cite{wang2026earlytom} & CVPR'26 & 25\% & 61.9 & \underline{57.3} & 65.0 & \underline{59.1} & \underline{55.3} & 59.7 & 96.5 \\
AOT~\cite{li2026token} & CVPR'26 & 25\% & \underline{62.4} & 56.2 & \textbf{66.2} & 58.8 & \textbf{55.4} & \underline{59.8} & \underline{96.6} \\
\rowcolor[RGB]{216,229,223}
\textbf{SimpleComp} & -- & 25\% & \textbf{62.7} & \textbf{58.1} & \underline{65.9} & \textbf{60.5} & 54.8 & \textbf{60.4} & \textbf{97.6} \\
\midrule
VisionZip~\cite{yang2025visionzip} & CVPR'25 & 20\% & \textbf{62.0} & \underline{57.0} & 65.3 & \underline{60.5} & \textbf{55.5} & \textbf{60.1} & \textbf{97.1} \\
EarlyTom~\cite{wang2026earlytom} & CVPR'26 & 20\% & 61.2 & 54.3 & 63.9 & 59.1 & 54.4 & 58.6 & 94.7 \\
AOT~\cite{li2026token} & CVPR'26 & 20\% & \textbf{62.0} & 56.4 & \underline{65.7} & \textbf{61.0} & \underline{55.1} & \underline{60.0} & \textbf{97.1} \\
\rowcolor[RGB]{216,229,223}
\textbf{SimpleComp} & -- & 20\% & \underline{61.6} & \textbf{57.7} & \textbf{65.8} & 60.3 & 54.1 & 59.9 & \underline{96.8} \\
\bottomrule
\end{tabular}%
}
\end{table*}

%% file: tabs/sup_internvl3.tex
\begin{table*}[!ht]
\caption{Comparison with state-of-the-art methods on InternVL3-8B. ``Avg. Score'' and ``Avg. \%'' denote the average score across benchmarks and the relative performance to the unpruned model. The best and second-best results are in \textbf{bold} and \underline{underlined}, respectively.}
\label{tab:sup_internvl3}
\centering
\resizebox{\textwidth}{!}{%
\begin{tabular}{l|c|c|ccccc|cc}
\toprule
\multirow{2}{*}{Method}
& \multirow{2}{*}{Source}
& \multirow{2}{*}{\shortstack{Before LLM\\Retained Ratio}}
& \multirow{2}{*}{\shortstack{VideoMME\\$\uparrow$}}
& \multirow{2}{*}{\shortstack{LongVideoBench\\$\uparrow$}}
& \multirow{2}{*}{\shortstack{MLVU\\$\uparrow$}}
& \multirow{2}{*}{\shortstack{MVBench\\$\uparrow$}}
& \multirow{2}{*}{\shortstack{EgoSchema\\$\uparrow$}}
& \multicolumn{2}{|c}{Avg. $\uparrow$} \\
& & & & & & & & Score & \% \\
\midrule
\rowcolor[RGB]{235,235,235}
InternVL3-8B~\cite{zhu2025internvl3} & -- & 100\% & 64.9 & 58.7 & 68.8 & 74.1 & 62.6 & 65.8 & 100.0 \\
\midrule
VisionZip~\cite{yang2025visionzip} & CVPR'25 & 25\% & \textbf{64.9} & \textbf{58.8} & \underline{68.2} & \textbf{74.3} & \underline{61.2} & \textbf{65.5} & \textbf{99.5} \\
EarlyTom~\cite{wang2026earlytom} & CVPR'26 & 25\% & \underline{64.3} & \underline{58.5} & \textbf{68.4} & \underline{73.9} & \textbf{61.3} & \underline{65.3} & \underline{99.2} \\
AOT~\cite{li2026token} & CVPR'26 & 25\% & 61.8 & 55.0 & 65.2 & 73.7 & 59.3 & 63.0 & 95.7 \\
\rowcolor[RGB]{216,229,223}
\textbf{SimpleComp} & -- & 25\% & 64.2 & 58.2 & 68.0 & 72.6 & 60.7 & 64.7 & 98.4 \\
\midrule
VisionZip~\cite{yang2025visionzip} & CVPR'25 & 20\% & \textbf{64.1} & 56.7 & \underline{67.8} & \textbf{74.0} & \textbf{61.0} & \underline{64.7} & \underline{98.4} \\
EarlyTom~\cite{wang2026earlytom} & CVPR'26 & 20\% & 63.5 & \textbf{57.7} & \textbf{68.1} & \underline{73.8} & \textbf{61.0} & \textbf{64.8} & \textbf{98.5} \\
AOT~\cite{li2026token} & CVPR'26 & 20\% & 61.4 & 55.1 & 64.2 & 73.0 & 58.7 & 62.5 & 94.9 \\
\rowcolor[RGB]{216,229,223}
\textbf{SimpleComp} & -- & 20\% & \underline{64.0} & \underline{57.3} & \underline{67.8} & 72.1 & \underline{60.2} & 64.3 & 97.7 \\
\bottomrule
\end{tabular}%
}
\end{table*}    

%% file: tabs/sup_diff_baselines.tex
\begin{table*}[t]
\caption{Comparison with simple baselines on LLaVA-OneVision-7B. ``Avg. Score'' and ``Avg. \%'' denote the average score across benchmarks and the relative performance to the unpruned model. The best and second-best results are in \textbf{bold} and \underline{underlined}, respectively.}
\label{tab:sup_diff_baselines}
\centering
\resizebox{\textwidth}{!}{%
\begin{tabular}{l|c|ccccc|cc}
\toprule
\multirow{2}{*}{Method}
& \multirow{2}{*}{\shortstack{Before LLM\\Retained Ratio}}
& \multirow{2}{*}{\shortstack{VideoMME\\$\uparrow$}}
& \multirow{2}{*}{\shortstack{LongVideoBench\\$\uparrow$}}
& \multirow{2}{*}{\shortstack{MLVU\\$\uparrow$}}
& \multirow{2}{*}{\shortstack{MVBench\\$\uparrow$}}
& \multirow{2}{*}{\shortstack{EgoSchema\\$\uparrow$}}
& \multicolumn{2}{|c}{Avg. $\uparrow$} \\
& & & & & & & Score & \% \\
\midrule
\rowcolor[RGB]{235,235,235}
LLaVA-OV-7B~\cite{li2025llava}
& 100\% & 58.6 & 56.4 & 63.0 & 58.3 & 60.4 & 59.3 & 100.0 \\
\midrule
Spatial Average Pooling
& 25\% & 56.7 & 55.3 & \underline{62.4} & 56.9 & \underline{59.9} & \underline{58.2} & \underline{98.1} \\
Spatiotemporal Average Pooling
& 25\% & 56.6 & \underline{56.2} & 61.0 & 57.2 & 58.8 & 57.9 & 97.7 \\
Spatial Grid Sampling
& 25\% & \underline{56.8} & 54.2 & 62.0 & 57.1 & 59.5 & 57.9 & 97.6 \\
Spatiotemporal Grid Sampling
& 25\% & 56.2 & 55.0 & 60.2 & \underline{57.6} & 59.3 & 57.7 & 97.2 \\
\rowcolor[RGB]{216,229,223}
\textbf{SimpleComp}
& 25\% & \textbf{58.4} & \textbf{57.6} & \textbf{63.8} & \textbf{58.1} & \textbf{60.4} & \textbf{59.6} & \textbf{100.5} \\
\midrule
Spatial Average Pooling
& 20\% & 55.8 & \underline{56.2} & \underline{62.4} & 55.8 & 58.4 & 57.7 & 97.3 \\
Spatiotemporal Average Pooling
& 20\% & 55.5 & 54.2 & 59.6 & \underline{56.5} & 58.9 & 56.9 & 96.0 \\
Spatial Grid Sampling
& 20\% & \underline{57.4} & 54.7 & 61.8 & \underline{56.5} & 58.7 & \underline{57.8} & \underline{97.4} \\
Spatiotemporal Grid Sampling
& 20\% & 55.7 & 54.7 & 61.4 & \underline{56.5} & \underline{59.1} & 57.5 & 96.9 \\
\rowcolor[RGB]{216,229,223}
\textbf{SimpleComp}
& 20\% & \textbf{57.6} & \textbf{56.6} & \textbf{62.5} & \textbf{57.0} & \textbf{59.8} & \textbf{58.7} & \textbf{98.9} \\
\midrule
Spatial Average Pooling
& 15\% & 55.9 & 55.0 & 60.1 & 55.1 & 58.8 & 57.0 & 96.0 \\
Spatiotemporal Average Pooling
& 15\% & 54.6 & \underline{55.1} & 59.9 & 55.8 & 58.5 & 56.8 & 95.7 \\
Spatial Grid Sampling
& 15\% & \underline{56.7} & 54.1 & \textbf{62.0} & 55.2 & \underline{59.1} & \underline{57.4} & \underline{96.7} \\
Spatiotemporal Grid Sampling
& 15\% & 54.9 & 54.4 & 60.0 & \underline{56.8} & 58.8 & 57.0 & 96.0 \\
\rowcolor[RGB]{216,229,223}
\textbf{SimpleComp}
& 15\% & \textbf{57.3} & \textbf{56.9} & \underline{61.8} & \textbf{57.4} & \textbf{60.1} & \textbf{58.7} & \textbf{98.9} \\
\bottomrule
\end{tabular}
}
\end{table*}

%% file: tabs/sup_internvl3_2b.tex
\begin{table*}[!ht]
\caption{Comparison with state-of-the-art methods on InternVL3-2B. ``Avg. Score'' and ``Avg. \%'' denote the average score across benchmarks and the relative performance to the unpruned model. AOT is omitted at 1\% because its structural constraints require more tokens than the available budget. The best and second-best results are in \textbf{bold} and \underline{underlined}, respectively.}
\label{tab:internvl3_2b}
\centering
\resizebox{\textwidth}{!}{%
\begin{tabular}{l|c|c|ccccc|cc}
\toprule
\multirow{2}{*}{Method}
& \multirow{2}{*}{Source}
& \multirow{2}{*}{\shortstack{Before LLM\\Retained Ratio}}
& \multirow{2}{*}{\shortstack{VideoMME\\$\uparrow$}}
& \multirow{2}{*}{\shortstack{LongVideoBench\\$\uparrow$}}
& \multirow{2}{*}{\shortstack{MLVU\\$\uparrow$}}
& \multirow{2}{*}{\shortstack{MVBench\\$\uparrow$}}
& \multirow{2}{*}{\shortstack{EgoSchema\\$\uparrow$}}
& \multicolumn{2}{|c}{Avg. $\uparrow$} \\
& & & & & & & & Score & \% \\
\midrule
\rowcolor[RGB]{235,235,235}
InternVL3-2B~\cite{zhu2025internvl3}
& -- & 100\% & 57.2 & 53.7 & 63.2 & 68.6 & 49.1 & 58.4 & 100.0 \\
\midrule
VisionZip~\cite{yang2025visionzip}
& CVPR'25 & 25\%
& \underline{56.9} & \underline{53.3} & \underline{63.7}
& \textbf{67.3} & \textbf{47.9}
& \underline{57.8} & \underline{99.1} \\
EarlyTom~\cite{wang2026earlytom}
& CVPR'26 & 25\%
& 55.6 & 51.5 & 63.6 & 66.9 & 47.5 & 57.1 & 97.8 \\
AOT~\cite{li2026token}
& CVPR'26 & 25\%
& 50.9 & 49.6 & 60.3 & 65.6 & 45.1 & 54.3 & 93.0 \\
\rowcolor[RGB]{216,229,223}
\textbf{SimpleComp}
& -- & 25\%
& \textbf{57.8} & \textbf{53.9} & \textbf{63.8}
& \underline{67.1} & \underline{47.6}
& \textbf{58.0} & \textbf{99.4} \\
\midrule
VisionZip~\cite{yang2025visionzip}
& CVPR'25 & 20\%
& \underline{57.0} & \underline{52.5} & 62.7
& 65.2 & \textbf{47.4}
& \underline{56.9} & \underline{97.6} \\
EarlyTom~\cite{wang2026earlytom}
& CVPR'26 & 20\%
& 54.9 & 50.7 & \underline{63.2}
& \underline{65.7} & 47.1 & 56.3 & 96.5 \\
AOT~\cite{li2026token}
& CVPR'26 & 20\%
& 51.3 & 49.8 & 59.8 & 65.6 & 44.5 & 54.2 & 92.9 \\
\rowcolor[RGB]{216,229,223}
\textbf{SimpleComp}
& -- & 20\%
& \textbf{57.4} & \textbf{53.4} & \textbf{63.5}
& \textbf{66.7} & \underline{47.2}
& \textbf{57.6} & \textbf{98.8} \\
\midrule
VisionZip~\cite{yang2025visionzip}
& CVPR'25 & 15\%
& \underline{56.2} & \underline{52.1} & 62.3
& 63.2 & \underline{46.3}
& \underline{56.0} & \underline{96.0} \\
EarlyTom~\cite{wang2026earlytom}
& CVPR'26 & 15\%
& 54.5 & 50.9 & \underline{62.4}
& 64.3 & 46.0 & 55.6 & 95.3 \\
AOT~\cite{li2026token}
& CVPR'26 & 15\%
& 52.0 & 49.4 & 59.2
& \underline{65.3} & 43.9 & 53.9 & 92.4 \\
\rowcolor[RGB]{216,229,223}
\textbf{SimpleComp}
& -- & 15\%
& \textbf{57.7} & \textbf{53.0} & \textbf{63.6}
& \textbf{66.5} & \textbf{46.4}
& \textbf{57.4} & \textbf{98.4} \\
\midrule
VisionZip~\cite{yang2025visionzip}
& CVPR'25 & 10\%
& \underline{54.6} & \underline{51.2} & \underline{61.5}
& 59.9 & \underline{45.0}
& \underline{54.4} & \underline{93.3} \\
EarlyTom~\cite{wang2026earlytom}
& CVPR'26 & 10\%
& 53.7 & 49.9 & 60.1 & 62.8 & 44.3 & 54.2 & 92.8 \\
AOT~\cite{li2026token}
& CVPR'26 & 10\%
& 51.3 & 48.1 & 59.4
& \underline{63.4} & 41.6 & 52.7 & 90.4 \\
\rowcolor[RGB]{216,229,223}
\textbf{SimpleComp}
& -- & 10\%
& \textbf{56.4} & \textbf{53.3} & \textbf{62.4}
& \textbf{65.4} & \textbf{45.4}
& \textbf{56.6} & \textbf{97.0} \\
\midrule
VisionZip~\cite{yang2025visionzip}
& CVPR'25 & 5\%
& \underline{52.6} & \underline{48.9} & 57.7
& 54.6 & 41.2 & 51.0 & 87.4 \\
EarlyTom~\cite{wang2026earlytom}
& CVPR'26 & 5\%
& 51.3 & 47.9 & \underline{60.1}
& 58.8 & \underline{41.8}
& \underline{52.0} & \underline{89.0} \\
AOT~\cite{li2026token}
& CVPR'26 & 5\%
& 49.4 & 47.9 & 57.1
& \underline{60.1} & 39.2 & 50.7 & 86.9 \\
\rowcolor[RGB]{216,229,223}
\textbf{SimpleComp}
& -- & 5\%
& \textbf{56.0} & \textbf{52.5} & \textbf{61.5}
& \textbf{64.3} & \textbf{42.0}
& \textbf{55.3} & \textbf{94.7} \\
\midrule
VisionZip~\cite{yang2025visionzip}
& CVPR'25 & 1\%
& 44.1 & 43.1 & 51.6 & 44.0 & 30.2 & 42.6 & 73.0 \\
EarlyTom~\cite{wang2026earlytom}
& CVPR'26 & 1\%
& \underline{46.4} & \underline{45.2} & \underline{54.3}
& \underline{51.5} & \underline{34.0}
& \underline{46.3} & \underline{79.3} \\
\rowcolor[RGB]{216,229,223}
\textbf{SimpleComp}
& -- & 1\%
& \textbf{52.6} & \textbf{48.5} & \textbf{56.6}
& \textbf{59.4} & \textbf{36.0}
& \textbf{50.6} & \textbf{86.8} \\
\bottomrule
\end{tabular}%
}
\end{table*}

%% file: tabs/sup_runtime.tex
\begin{table*}[t]
\centering
\begin{minipage}[t]{0.49\textwidth}
\centering
\caption{Runtime breakdown on LLaVA-OneVision-7B.}
\label{tab:sup_runtime_llava_ov}
\resizebox{\linewidth}{!}{%
\begin{tabular}{l|c|ccc}
\toprule
Method
& \shortstack{Retained\\Ratio}
& \shortstack{Clustering Core\\(ms) $\downarrow$}
& \shortstack{Compression\\(ms) $\downarrow$}
& \shortstack{LLM Prefill\\(ms) $\downarrow$} \\
\midrule
\rowcolor[RGB]{235,235,235}
LLaVA-OV-7B~\cite{li2025llava}
& 100\% & -- & -- & 559.0 \\
\midrule
AOT~\cite{li2026token}
& 25\% & -- & 301.0 & 144.2 \\
\rowcolor[RGB]{216,229,223}
\textbf{SimpleComp}
& 25\% & 152.6 & 155.5 & 144.8 \\
\midrule
AOT~\cite{li2026token}
& 20\% & -- & 273.1 & 116.8 \\
\rowcolor[RGB]{216,229,223}
\textbf{SimpleComp}
& 20\% & 128.5 & 131.3 & 121.5 \\
\midrule
AOT~\cite{li2026token}
& 15\% & -- & 251.8 & 93.2 \\
\rowcolor[RGB]{216,229,223}
\textbf{SimpleComp}
& 15\% & 98.6 & 101.4 & 93.0 \\
\midrule
AOT~\cite{li2026token}
& 10\% & -- & 211.7 & 70.7 \\
\rowcolor[RGB]{216,229,223}
\textbf{SimpleComp}
& 10\% & 65.5 & 68.3 & 70.4 \\
\midrule
AOT~\cite{li2026token}
& 5\% & -- & 173.0 & 47.4 \\
\rowcolor[RGB]{216,229,223}
\textbf{SimpleComp}
& 5\% & 36.4 & 39.1 & 46.8 \\
\bottomrule
\end{tabular}%
}
\end{minipage}
\hfill
\begin{minipage}[t]{0.49\textwidth}
\centering
\caption{Runtime breakdown on LLaVA-Video-7B.}
\label{tab:sup_runtime_llava_video}
\resizebox{\linewidth}{!}{%
\begin{tabular}{l|c|ccc}
\toprule
Method
& \shortstack{Retained\\Ratio}
& \shortstack{Clustering Core\\(ms) $\downarrow$}
& \shortstack{Compression\\(ms) $\downarrow$}
& \shortstack{LLM Prefill\\(ms) $\downarrow$} \\
\midrule
\rowcolor[RGB]{235,235,235}
LLaVA-Video-7B~\cite{zhang2025llava}
& 100\% & -- & -- & 1076.4 \\
\midrule
AOT~\cite{li2026token}
& 25\% & -- & 448.7 & 311.4 \\
\rowcolor[RGB]{216,229,223}
\textbf{SimpleComp}
& 25\% & 374.8 & 461.1 & 310.1 \\
\midrule
AOT~\cite{li2026token}
& 20\% & -- & 410.8 & 265.9 \\
\rowcolor[RGB]{216,229,223}
\textbf{SimpleComp}
& 20\% & 320.5 & 404.0 & 264.7 \\
\midrule
AOT~\cite{li2026token}
& 15\% & -- & 350.2 & 219.2 \\
\rowcolor[RGB]{216,229,223}
\textbf{SimpleComp}
& 15\% & 243.0 & 324.7 & 218.9 \\
\midrule
AOT~\cite{li2026token}
& 10\% & -- & 285.7 & 164.3 \\
\rowcolor[RGB]{216,229,223}
\textbf{SimpleComp}
& 10\% & 164.2 & 238.2 & 174.9 \\
\midrule
AOT~\cite{li2026token}
& 5\% & -- & 217.4 & 123.7 \\
\rowcolor[RGB]{216,229,223}
\textbf{SimpleComp}
& 5\% & 84.1 & 148.5 & 123.7 \\
\bottomrule
\end{tabular}%
}
\end{minipage}
\end{table*}